\documentclass[twoside,jmlr,11pt]{article}

\usepackage{blindtext}

\usepackage[preprint]{jmlr2e}
\usepackage{booktabs}
\usepackage{amsmath}


\usepackage{lastpage}
\ShortHeadings{Trusted Federated Learning at Edge-Scale}{Onobhayedo Oamen}
\firstpageno{1}

\begin{document}

\title{Trustless Federated Learning at Edge-Scale: A Compositional Architecture
	for Decentralized, Verifiable, and Incentive-Aligned Coordination}

\author{\name Pius Onobhayedo \email pius.onobhayedo@usc.edu \\
       \addr Marshall School of Business\\
       University of Southern California\\
       Los Angeles, CA 90007, USA
       \AND
       \name Paul Osemudiame Oamen \email p.oamen.25@abdn.ac.uk \\
       \addr School of Natural and Computing Sciences\\
       University of Aberdeen\\
       King's College\\
       Aberdeen\\
       AB24 3FX, UK
}

\editor{}

\maketitle

\begin{abstract}%   <- trailing '%' for backward compatibility of .sty file
Artificial intelligence is retracing the Internet's path from centralized provision to distributed creation. Initially, resource-intensive computation concentrates within institutions capable of training and serving large models. Eventually, as federated learning matures, billions of edge devices holding sensitive data will be able to collectively improve models without surrendering raw information, enabling both contribution and consumption at scale.

This democratic vision remains unrealized due to certain compositional gaps; aggregators handle updates without accountability, economic mechanisms are lacking and even when present remain vulnerable to gaming, coordination serializes state modifications limiting scalability, and governance permits retroactive manipulation.
This work addresses these gaps by leveraging cryptographic receipts to prove aggregation correctness, geometric novelty measurement to prevent incentive gaming, parallel object ownership to achieve linear scalability, and time-locked policies to check retroactive manipulation.

The product of this work is a design architecture—not an actual implementation—that seeks to pass the baton in the race toward truly collaborative intelligence; an intelligence of the people, by the people, for the people.
\end{abstract}

\begin{keywords}
  federated learning, edge-scale, trustless, verifiable, incentivized, pgot architecture
\end{keywords}

\section{Introduction}

\subsection{Motivation: The Trust Gap in Edge Intelligence}

\begin{figure}[ht]           % t = top, b = bottom, p = page of floats; use [htbp] if you prefer
	\centering
	\includegraphics[width=0.85\textwidth]{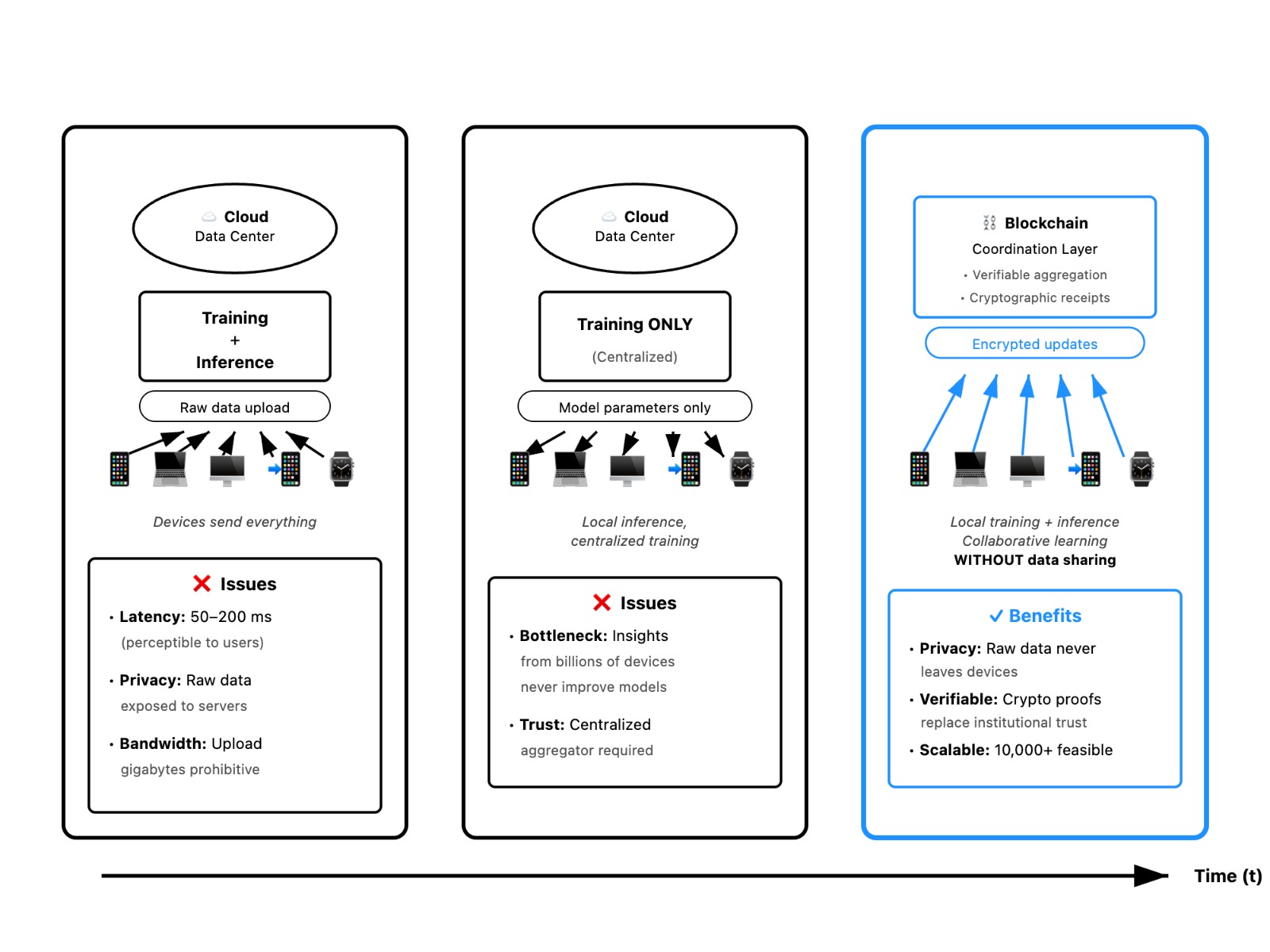}   % pdf, png, jpg all work
	\caption{Evolution of AI architectural paradigms: from centralized cloud training, to distributed inference with centralized training, to blockchain-coordinated edge-native collaborative learning.}
	\label{fig:architecture_evolution}        % use this label to cite with \ref{fig:yourlabel}
\end{figure}

Artificial intelligence stands at the threshold of its third architectural era. The first era centralized computation in data centers, trading latency and privacy for scale. The second distributed inference to the edge but kept training centralized, creating a bottleneck where insights from billions of devices never improved the models serving them. The field now envisions a third paradigm: edge-native intelligence, where learning happens continuously, locally, and collaboratively without surrendering raw data to distant servers
(Figure~\ref{fig:architecture_evolution}).

Consider smartphone autocomplete. Cloud-based prediction introduces 50 to 200 milliseconds of latency, perceptible enough to frustrate users \citep{Nielsen1993}. Pure on-device models plateau quickly, unable to predict jargon or patterns they have never encountered. Federated Learning (FL) \citep{McMahan2017} promised to bridge this gap by enabling devices to collaboratively train shared models through encrypted parameter updates. Early deployments including Google's Gboard \citep{Hard2018} and Apple's QuickType \citep{Apple2017} demonstrated billion-user scale success, proving that cross-device learning could work in practice.

Yet these successes relied on institutional trust: users trusted aggregators to handle updates responsibly, apply privacy protections correctly, and make fair decisions about model quality. That trust is eroding. High-profile incidents from Cambridge Analytica \citep{Cadwalladr2018} to algorithmic bias \citep{Buolamwini2018} and opaque content moderation \citep{Gillespie2018} have made users, regulators, and enterprises demand verifiable guarantees rather than corporate promises. In federated systems, the aggregator becomes a single point of failure, deciding whose updates to accept, how to weight them, and whether safety standards are met. When trust breaks, there is no cryptographic receipt, no audit trail, no recourse.

Blockchain technology offers a complementary accountability layer. By replacing trusted intermediaries with verifiable protocols, blockchains enable trustless payments \citep{Nakamoto2008}, decentralized file storage \citep{Benet2014}, and transparent supply chains \citep{Crosby2016}. The insight maps naturally to FL's governance challenge: a system should prove, after the fact, that a model was trained under specific rules, by specific contributors, with specific privacy guarantees, all without revealing the training data itself.

\subsection{The Composition Gap in Blockchain-Federated Learning}

Blockchain-federated learning (BC-FL) remains embryonic despite five years of research. A 2021 survey identified 47 foundational proposals \citep{Nguyen2021}; a 2024 update reviewing 135 papers documents exponential growth \citep{Ning2024}. Systems like BlockFLA, FLChain, and BLADE-FL have progressed to simulated scenarios in healthcare and IoT security. Yet fewer than five systems have moved beyond proof-of-concept to sustained operation at meaningful scale \citep{Ning2024}.

The barrier is not technical immaturity of individual components. Secure aggregation (SecAgg) \citep{Bonawitz2017} provides dropout-tolerant privacy. Zero-knowledge proofs enable verifiable aggregation \citep{wang2025zkFL}. Differential privacy bounds statistical inference \citep{Abadi2016}. Time-locked governance prevents retroactive manipulation \citep{CompoundFinance2020, Laurie2013}. Byzantine-robust aggregation filters adversarial updates through coordinate-wise median and trimmed mean operations, achieving order-optimal statistical rates \citep{Yin2018}. However, recent analysis reveals counter-intuitive results: simple mean aggregation can outperform specialized robust aggregators under certain attack models, particularly label poisoning on heterogeneous data \citep{Peng2024}, suggesting robustness mechanisms should be attack-specific rather than universally conservative. The challenge is compositional: these mechanisms exhibit emergent vulnerabilities at their boundaries that single-purpose designs never anticipated.

\begin{table}[ht]
	\centering
	\caption{Literature capabilities versus remaining composition gaps in blockchain-federated learning at edge scale.}
	\label{tab:composition-gap}
	\begin{tabular}{|p{3.5cm}|p{3.5cm}|p{3.5cm}|p{3.5cm}|}
		\toprule
		Challenge & Literature Delivers & Composition Gap & Our Mechanism \\
		\midrule
		Verifiable Aggregation & zkFL: ZK proofs for unweighted sums; SecAgg: dropout tolerance & No system composes weighted aggregation, dropout recovery, Byzantine fallback in single audit trail & Proof-carrying aggregation \\
		\hline
		Incentive Alignment & Utility-based rewards; contribution evaluation & No system resists replay and sybil attacks through directional novelty measurement & Geometric novelty decomposition \\
		\hline
		Scalable Coordination & DAG-based parallel execution; hierarchical sharding & No BC-FL system maps owned registries and shared state with validated 10,000+ scale costs & Object-centric parallelism \\
		\hline
		Policy Governance & Time-locks; Certificate Transparency & No system binds complete policy bundles to rounds preventing retroactive manipulation & Time-locked governance \\
		\bottomrule
	\end{tabular}
\end{table}

Table~\ref{tab:composition-gap} synthesizes this gap, showing what 2020-2025 literature delivers versus what remains unsolved when deploying at edge-scale serving 10,000 or more heterogeneous contributors.

Early BC-FL designs naively stored model weights on-chain, colliding with blockchain throughput limits (Bitcoin: 7 tx/sec \citep{Nakamoto2008}; Ethereum: 15-30 tx/sec \citep{Wood2014}).

Later proposals moved weights off-chain but encountered a subtler challenge: general-purpose blockchains optimize for financial transactions with sub-second finality and dynamic fee markets, fundamentally mismatched to FL's requirements of predictable costs over hours-long coordination. Without cost predictability, consumer-facing micro-fee models become infeasible.

\subsection{Our Contributions}

This paper introduces an architecture addressing the compositional gap through four mechanisms. Each solves a specific problem preventing BC-FL from moving beyond proof-of-concept.

\textbf{Proof-carrying aggregation} addresses how to prove aggregation correctness without revealing individual contributions. Traditional federated learning provides no mathematical proof, only operational logs that auditors cannot independently verify. Secure aggregation \citep{Bonawitz2017} hides individual updates through cryptographic masking but provides no integrity guarantees. An aggregator could forge results, selectively exclude contributors, or ignore declared privacy budgets. The problem compounds when devices drop out mid-round or adversarial updates require Byzantine filtering. Did the aggregator reconstruct dropouts correctly? Was the declared filtering rule applied consistently? We introduce SumIntegrityProofs that transform this black box into verifiable computation, producing cryptographic receipts binding outputs to weighted combinations of masked inputs, dropout reconstruction decisions, and Byzantine filtering rules.

\begin{figure}[ht]
	\centering
	\includegraphics[width=0.85\textwidth]{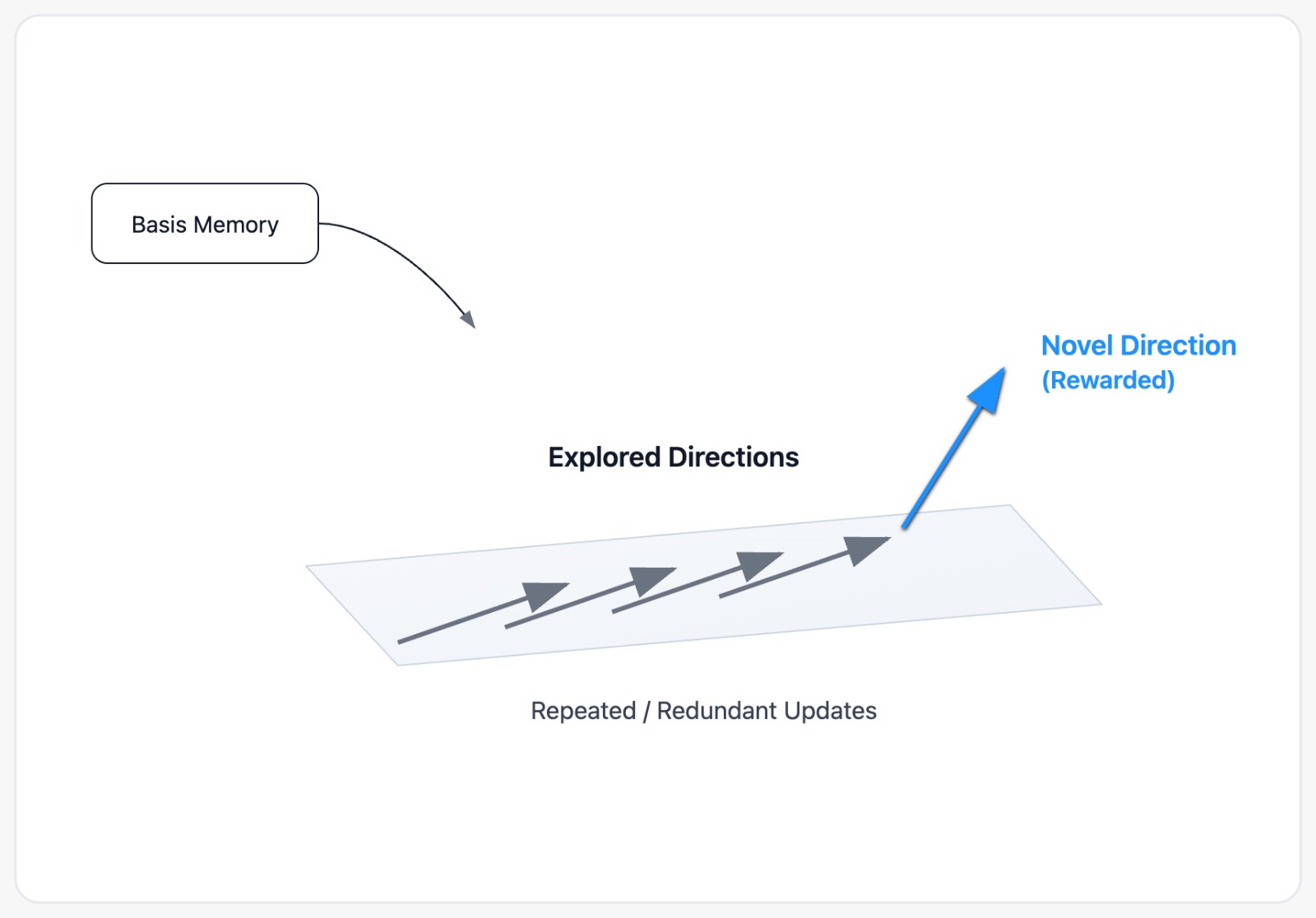}
	\caption{Geometric novelty decomposition for replay-resistant contribution measurement. Novel directions (perpendicular to basis memory) receive rewards, while repeated or redundant updates (parallel to explored directions) yield zero novelty score.}
	\label{fig:geometric_novelty}
\end{figure}

\textbf{Geometric novelty decomposition} addresses economic sustainability in trustless systems. Most BC-FL proposals assume altruistic participation, but running local training on smartphones consumes 5-20\% battery charge \citep{lai2022oort}, becoming prohibitive over hundreds of rounds without compensation. Yet compensation creates a deeper problem: how do you pay fairly when secure aggregation deliberately hides individual updates? Paying everyone equally rewards free-riders. Utility-based scoring is fundamentally gameable \citep{Xu2024}. Worse are replay attacks (resubmitting the same valuable update across rounds) and sybil attacks (splitting updates across fake identities to claim multiple rewards). Detecting these attacks requires storing all historical updates, violating privacy or becoming computationally prohibitive. We solve this through directional contribution measurement using geometric basis projection, where the system maintains a basis of already-explored directions and rewards only perpendicular components representing genuine exploration (Figure~\ref{fig:geometric_novelty}).

\begin{figure}[ht]
	\centering
	\includegraphics[width=0.85\textwidth]{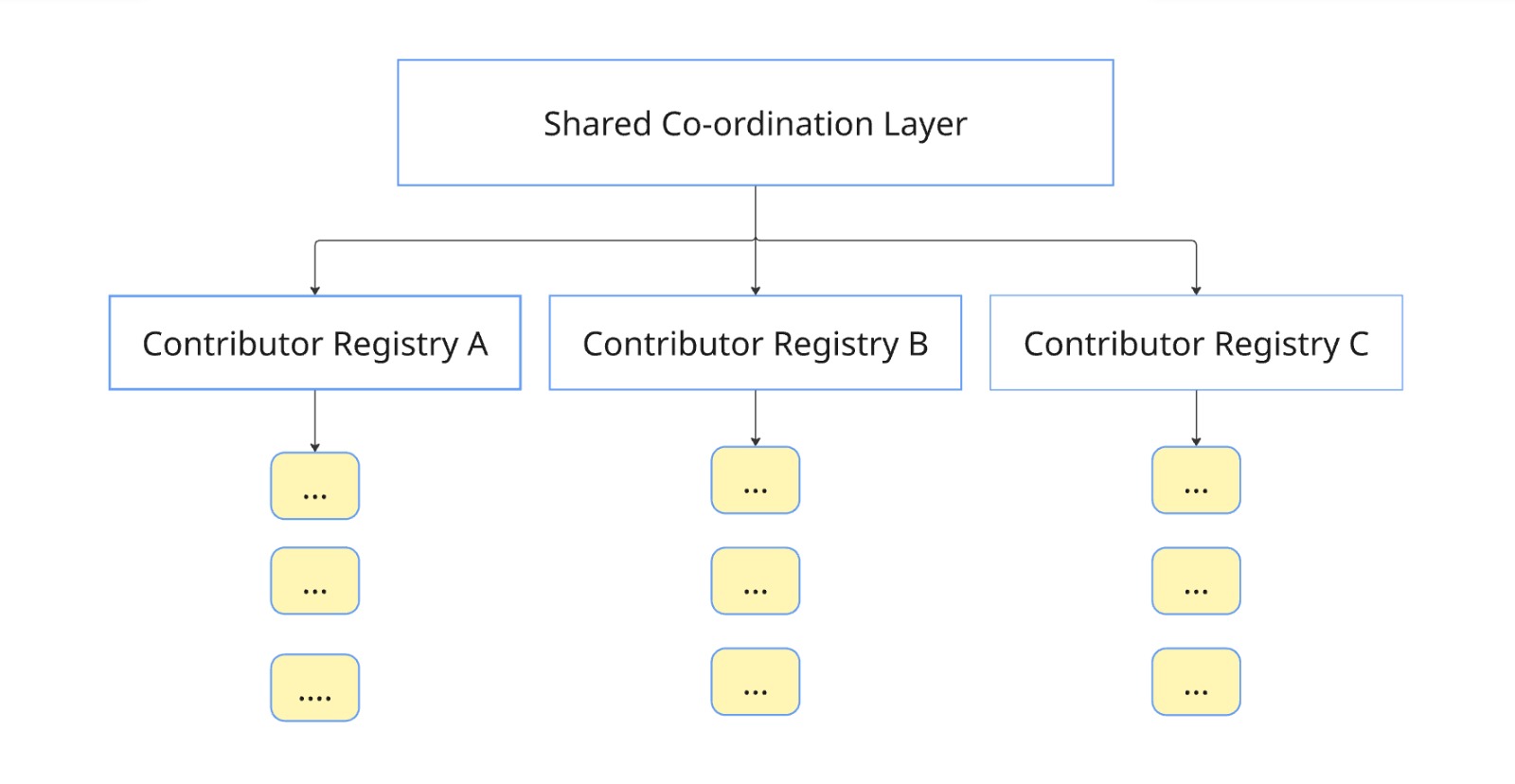}
	\caption{Object-centric coordination architecture decomposing state into parallel owned ContributorRegistry objects and infrequent shared coordination layer, enabling $O(N)$ scaling versus traditional $O(N^2)$ serialization.}
	\label{fig:object_coordination}
\end{figure}

\textbf{Object-centric coordination} solves the scalability bottleneck that general-purpose blockchains create. Traditional blockchain architectures serialize all state modifications through global consensus. When 10,000 contributors simultaneously update their reputation scores, privacy budgets, and participation logs, existing systems process these updates sequentially. This serialization is unnecessary: contributor A's reputation update has no logical dependency on contributor B's privacy accounting. We leverage Directed Acyclic Graph (DAG)-based Byzantine consensus \citep{danezis2022narwhal, blackshear2024mysticeti} with parallel state updates, decomposing coordination into independently-writable contributor registries and infrequent shared-object coordination (round configuration, model lineage, policy versions). This transforms O(N²) coordination complexity to O(N) parallel execution (Figure~\ref{fig:object_coordination}).

The architecture is designed for deployment on a purpose-built blockchain with fixed gas pricing and governance aligned to federated learning requirements.

\textbf{Time-locked governance} prevents retroactive rule manipulation. In traditional FL, aggregators adjust safety thresholds, privacy budgets, or admission criteria after observing training outcomes. A model shows unexpected bias? Lower the fairness threshold retroactively. A round consumed too much privacy budget? Increase the declared epsilon ex post facto. Simply recording rules on-chain proves insufficient if rules can change arbitrarily. An aggregator could propose a new policy in round 100, activate it immediately, and apply it retroactively to rounds 95-99. We introduce PolicyOracles with activation snapshots that make retroactive manipulation cryptographically provable fraud, through mandatory lock periods between proposal and activation.

These four mechanisms compose correctly to create a system where trust is verified through cryptographic proofs, where incentives align through geometric decomposition, where scale is achieved through parallel object updates, and where rules bind operators as firmly as participants. Security analysis demonstrates compositional correctness by reduction to established primitives: SecAgg \citep{Bonawitz2017}, differential privacy \citep{Abadi2016}, and Byzantine fault-tolerant consensus \citep{Castro1999}. Cost analysis confirms economic viability at \$0.001 per round for 10,000 contributors with 20M parameters.

This work is positioned as a design proposal demonstrating architectural feasibility through rigorous analysis, consistent with precedent from Dynamo \citep{decandia2007dynamo}, MapReduce \citep{dean2004mapreduce}, and DAG-based execution models \citep{danezis2022narwhal}. Full validation through testnet deployment remains important follow-on work, explicitly acknowledged as the primary limitation. We tag our proposed architecture with the acronym PGOT in line with the first letters of the four mechanisms.

\section{Background and Problem Context}

\subsection{Federated Learning and Privacy Primitives}

Federated learning emerged in 2017 as a paradigm enabling collaborative model training without centralizing raw data \citep{McMahan2017}. In the canonical cross-device setting, thousands to millions of mobile devices coordinate to improve a shared model: an aggregator broadcasts the current model, selected devices train locally, devices upload encrypted parameter updates, and the aggregator combines these into an improved global model. The approach addresses privacy (raw data never leaves devices), compliance (data residency regulations satisfied through on-device training), and practicality (uploading gigabytes of raw data is bandwidth-prohibitive). Yet it introduces challenges: data heterogeneity from non-identically distributed (non-IID) systems destabilizes optimization \citep{Kairouz2021}, system heterogeneity creates stragglers, and communication efficiency becomes critical on mobile uplinks.

Three privacy primitives form the foundation for trustless federated learning. Secure aggregation \citep{Bonawitz2017} ensures aggregators learn only the sum of client updates through cryptographic masking, where pairwise masks cancel when summed and dropout recovery reconstructs missing masks via Shamir's secret sharing. The protocol provides information-theoretic security but no proof of correct execution—malicious aggregators could forge results or selectively exclude contributions.

Differential privacy \citep{Abadi2016} bounds statistical inference through calibrated noise injection, with $\varepsilon$-DP guaranteeing nearly identical distributions whether any individual's data is included. The theoretical foundations of differential privacy reveal benefits beyond privacy alone: \cite{Wang2016} establish that $(\varepsilon,\delta)$-differential privacy implies algorithmic stability, preventing models from overfitting to individual contributors while providing generalization guarantees without explicit regularization. For practical implementations, \cite{Chaudhuri2011} demonstrate that objective perturbation—adding noise to the loss function before optimization—achieves better utility than output perturbation for strongly convex losses, with privacy costs scaling as $O(1/\sqrt{n})$ for $n$ contributors. However, aggregators claim specific epsilon values without publishing evidence that declared parameters were applied, preventing external verification.

Zero-knowledge federated learning \citep{Chen2022} constructs zk-SNARKs proving output aggregates equal committed client updates with compact 10--100~KB proofs, but targets only unweighted summation over fixed participant sets. Production systems require weighted aggregation, dropout tolerance, and Byzantine-robust fallback composed with cryptographic integrity proofs. These compositional requirements inform our architectural choice to apply differential privacy during aggregation rather than post hoc, leveraging both the privacy and stability properties that objective perturbation provides.

\subsection{Blockchain-Federated Learning: Progress and Gaps}

The BC-FL landscape has evolved from 47 foundational proposals in 2021 \citep{Nguyen2021} to 135 papers by 2024 \citep{Ning2024}, yet fewer than five systems have moved beyond proof-of-concept to sustained operation at meaningful scale. Early designs naively stored model weights on-chain, colliding with throughput constraints of 7-30 tx/sec. Three architectural patterns now dominate: hybrid storage separating model artifacts (IPFS) from on-chain cryptographic commitments, committee-based aggregation with staked validators coordinating off-chain while publishing receipts on-chain, and time-locked governance requiring proposals to undergo mandatory lock periods adapted from DeFi \citep{CompoundFinance2020} and Certificate Transparency \citep{Laurie2013}.

Recent systems show incremental progress. BlockFLA uses gradient similarity for contribution evaluation but relies on utility-based scoring vulnerable to manipulation \citep{Xu2024}. FLChain implements hierarchical sharding demonstrating 50-node deployments in simulation \citep{Bao2019FLChain, Majeed2019FLchain}. BLADE-FL proposes verifiable aggregation through committee signatures, but attestations endorse outcomes rather than procedures \citep{Li2021BLADEFL}. Three composition gaps persist: (1) custom versus public blockchain trade-off lacks empirical validation, with public chains offering mature security but facing gas volatility and governance misalignment, while custom chains enable fixed pricing but introduce maintenance burden; (2) no system composes weighted SecAgg, dropout recovery, and Byzantine fallback in unified cryptographic receipts; (3) time-locked proposals exist for single-parameter updates but not comprehensive bundles binding safety thresholds, privacy budgets, and admission criteria to training rounds.

\subsection{Incentive Mechanisms and Attacks}

Sustainable FL requires economic mechanisms aligning individual rationality with collective progress. Early BC-FL proposals adopted utility-based rewards paying contributors proportional to their impact on accuracy, with Shapley values providing theoretical foundation but requiring exponential computation. The ACE attack \citep{Xu2024} demonstrates that adversaries manipulate contribution evaluation by crafting updates appearing valuable in isolation while providing zero semantic improvement, reducing accuracy 20-40\% while earning 2-3× rewards. Reputation systems with stake decay offer alternatives but fail to address critical attacks. Replay attacks occur when contributors resubmit identical valuable updates across rounds, earning repeated rewards for singular effort. Detecting replay requires storing all historical updates, violating privacy or becoming computationally prohibitive. Sybil attacks occur when contributors split updates across fake identities to claim multiple base rewards, with existing systems unable to distinguish genuine collaboration from artificial fragmentation.

\section{THREAT MODEL AND ASSUMPTIONS}

\subsection{Adversary Classes}

Our architecture defends against five adversary classes exploiting different trust boundaries. \textbf{Curious aggregators} are honest-but-curious adversaries following protocol specifications but attempting to infer information about individual participants from aggregated data. Defense: secure aggregation hides individual updates through cryptographic masking with adapter-only transmission reducing gradient dimensionality, limiting gradient inversion attacks \citep{Zhu2019}. Privacy holds under fewer than \( t \) colluding committee members, where \( t = \lceil M/2 \rceil \) for \( M \) committee nodes.

\textbf{Byzantine contributors} submit malicious updates to poison model behavior, inject backdoors, or degrade accuracy \citep{Yin2018}. Defense: Byzantine-robust aggregation through statistical methods including coordinate-wise median filtering and trimmed mean estimation, with robust method selection bound to PolicyOracle configurations locked before round execution. We applied receipts document which robust rule, enabling verification of consistent enforcement.

\textbf{Sybil attackers} create multiple fake identities to claim repeated base rewards or amplify voting power. Defense: admission requirements including minimum stake deposits and device attestation raise identity creation costs, while geometric novelty decomposition measures contributions on aggregates rather than individuals. Splitting an update across 100 identities produces the identical aggregate novelty score, making identity splitting economically irrational.

\textbf{Governance manipulators} attempt to adjust system rules retroactively to benefit from hindsight. Defense: time-locked governance with mandatory lock periods between proposal and activation, where receipts cryptographically bind to policy versions active at round execution. Retroactive manipulation requires rewriting blockchain history, which Byzantine consensus prevents.

\textbf{Privacy inference attackers} combine information from multiple sources to reconstruct individual participant data or membership \citep{Shokri2017}. Defense: differential privacy bounds statistical inference, k-anonymity requirements prevent publication of cohort-level statistics when cohorts fall below size thresholds, and adapter-only transmission reduces dimensionality. Composition under heterogeneous data distributions remains an active research area \citep{Kairouz2021}.

\subsection{Security Assumptions}
Byzantine fault tolerance assumes at most f committee members exhibit Byzantine behavior, where $f < M/3$ for $M$ total nodes under standard BFT consensus \citep{Castro1999}. The illustrative 7-node committee supports $f=2$ Byzantine nodes; production systems should expand to M=20-100 nodes. Cryptographic primitives assume SHA-256 collision resistance, discrete logarithm hardness for Shamir secret sharing, and Pedersen commitment binding. The custom blockchain implements DAG-based Byzantine consensus with parallel execution for independent state updates and fixed gas pricing of approximately \$0.001 per round for 10,000 contributors. Off-chain storage maintains at least two geographically distributed replicas per artifact. Differential privacy accounting follows established methods \citep{Abadi2016} with default delta = $10^{-6}$.

\subsection{Out-of-Scope Threats}

Side-channel attacks including timing analysis and traffic pattern observation require deployment-level mitigations (submission jitter, traffic padding) inappropriate for protocol-level specification. Trusted execution environment (TEE) compromise affects client-side safety attestation, with the architecture providing defense-in-depth through committee-level safety gates as backstop. Adaptive adversarial machine learning where sophisticated attackers iteratively craft updates that pass safety filters yet subtly degrade quality receives partial mitigation through safety gates but requires ongoing red teaming rather than one-time solutions. Model extraction and intellectual property theft at inference time require additional defenses beyond this work's training-time guarantees.

\section{ARCHITECTURE OVERVIEW}

\subsection{Design Philosophy}

\begin{figure*}[ht]
	\centering
	\includegraphics[width=0.85\textwidth]{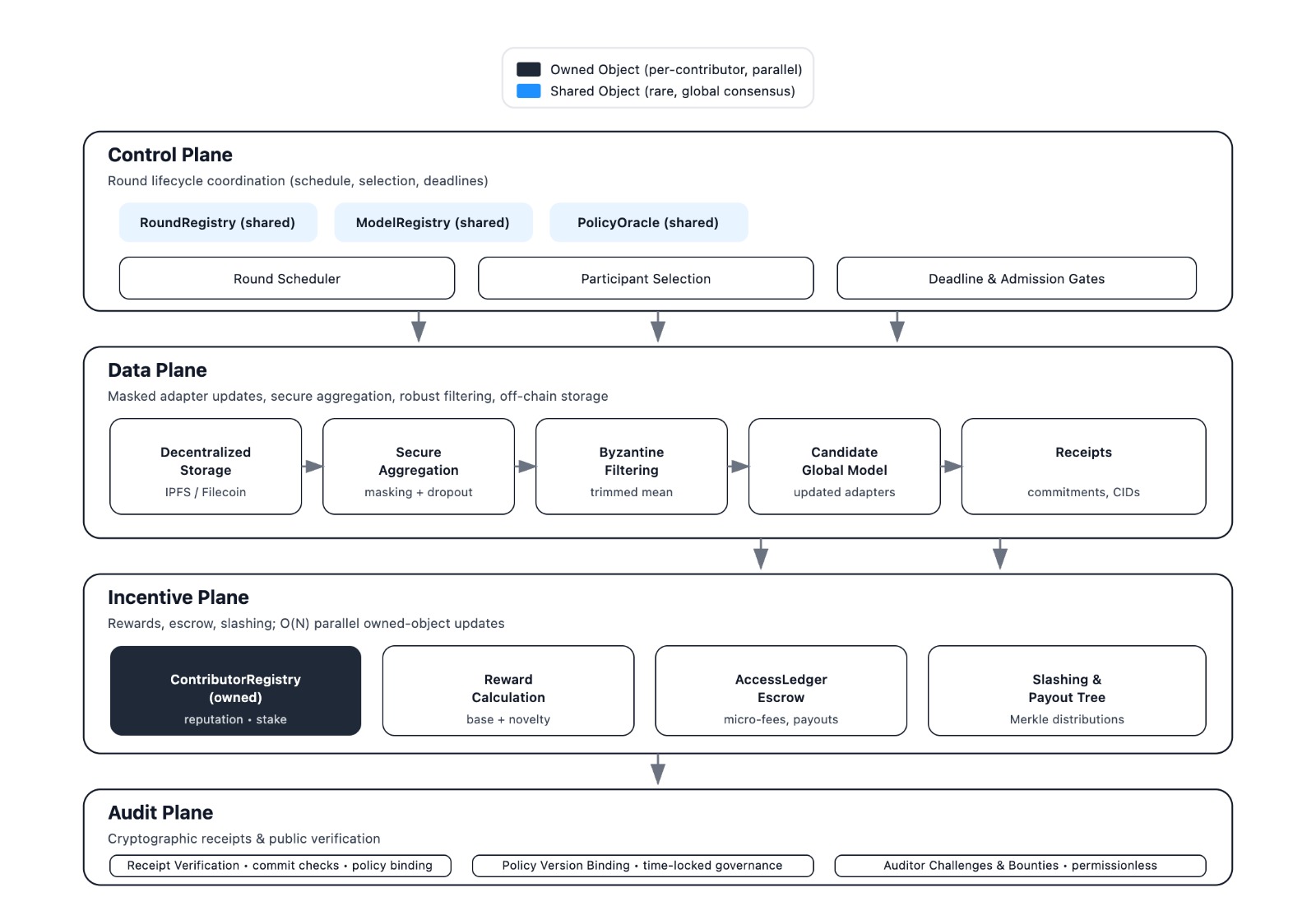}
	\caption{Separation of concerns across four operational planes: Control (round lifecycle), Data (secure aggregation), Incentive (reward distribution), and Audit (cryptographic verification). Owned objects (dark) enable parallel updates; shared objects (light) require consensus.}
	\label{fig:architecture_planes}
\end{figure*}

The architecture achieves compositional correctness through systematic separation of concerns across four operational planes: the control plane manages training lifecycle coordination (round scheduling, participant selection, deadline enforcement), the data plane handles model updates and secure aggregation with off-chain storage, the incentive plane calculates rewards and processes payments with safety-economics coupling, and the audit plane enables post-hoc verification through cryptographic receipts. This separation prevents failures from cascading: bugs in incentive calculation cannot corrupt model aggregation, and Byzantine committee members attempting reward manipulation produce invalid Merkle proofs that auditors detect immediately without affecting the training pipeline (Figure~\ref{fig:architecture_planes}).

\subsection{Participants and Object-Centric Coordination}

\begin{figure*}[t]
	\centering
	\includegraphics[width=0.9\textwidth]{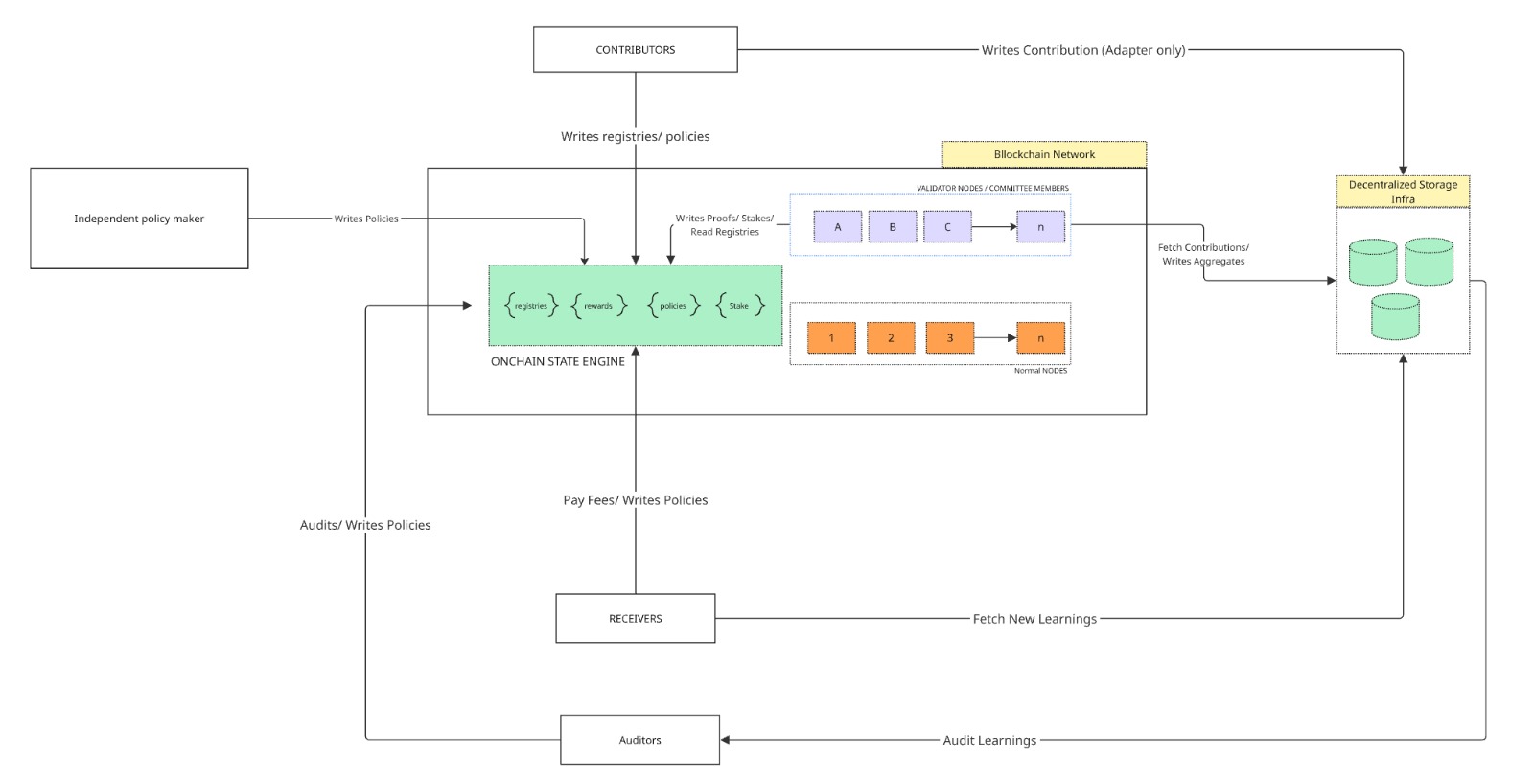}
	\caption{Complete system architecture showing six participant classes and their interactions. Contributors upload masked adapters to decentralized storage and submit CIDs to blockchain; committee validators fetch adapters, perform aggregation, and write results back to storage; receivers escrow fees and fetch trained models; auditors verify cryptographic receipts independently.}
	\label{fig:system_architecture}
\end{figure*}

Six participant classes interact through cryptographically enforced capabilities. Contributors perform local training and submit masked updates. Receivers fund training through escrowed micro-fees, paying only for successful rounds. Committee members coordinate aggregation and generate proofs, facing stake slashing for provable faults. PolicyOracle governs system rules through time-locked proposals. Auditors verify behavior using only public artifacts. Storage providers ensure durable availability through multi-provider replication (IPFS, Filecoin, Arweave). Table~\ref{tab:participant_capabilities} summarizes key capabilities and restrictions. Figure~\ref{fig:system_architecture} illustrates the complete system architecture, showing how these six participant classes interact through cryptographically enforced capabilities.

\begin{table}[ht]
	\centering
	\caption{Literature capabilities versus remaining composition gaps in blockchain-federated learning at edge scale.}
	\label{tab:participant_capabilities}
	\begin{tabular}{|p{3.5cm}|p{3.5cm}|p{3.5cm}|}
		\toprule
		Participant & Key Capabilities & Critical Restrictions  \\
		\midrule
		Contributors & Local training, masked submission, reward claims & Cannot access others' updates or modify shared state \\
		\hline
		Receivers & Escrow fees, download models, claim refunds & Cannot observe individual contributions or influence aggregation \\
		\hline
		Committee & Aggregate masked updates, generate proofs, vote in consensus & Cannot access plaintext updates or forge proofs undetected \\
		\hline
		PolicyOracle & Propose policies, activate after T\_lock & Cannot apply policies retroactively or bypass lock periods \\
		\hline
		Auditors & Verify receipts, file challenges, earn bounties & Cannot disrupt operations or access private data \\
		\hline
		Storage & Store artifacts, serve retrievals & Cannot modify stored data without detection \\
		\bottomrule
	\end{tabular}
\end{table}

The masked updates by contributors are published to a decentralized storage infrastructure (IPFS, Arweave), after which contributors submit content identifiers (CIDs) to the blockchain via the RoundRegistry smart contract. This architectural separation enables scalable coordination by decomposing storage responsibilities, ensuring that the blockchain maintains only lightweight CID references, while the storage layer handles large adapter artifacts. This design substantially reduces on-chain storage requirements compared to naive approaches that store adapters directly on-chain, enabling coordination at scale without blockchain state bloat. The committee members are made of all validator nodes who simultaneously serve three roles: blockchain validators (propose/attest blocks), smart contract executors (process on-chain transactions), and FL aggregators (combine masked updates). This unified role ensures accountability and eliminates coordination mismatches between consensus and aggregation. Validators fetch adapters from decentralized storage using CIDs during the aggregation phase. There are also normal nodes who simply store blockchain state and serve data via RPC endpoints. This enables permissionless verification with minimal resources, unlike validators which require high stake and uptime. Normal nodes can detect committee misbehavior through receipt verification but cannot disrupt operations.

The architecture achieves linear scaling by decomposing coordination into owned objects supporting parallel updates and shared objects requiring consensus. Owned ContributorRegistry objects contain reputation, stake, and privacy accounting, modified in parallel through single-owner authorization. Shared objects (RoundRegistry, ModelRegistry, PolicyOracle) maintain system-wide invariants but update rarely: round configuration changes every 2 hours, model lineage appends once per successful round, policy versions update weekly to monthly. This transforms O(N²) coordination to O(N) parallel execution for N contributors. Table~\ref{tab:object_decomposition} contrasts the decomposition.

\begin{table}[ht]
	\centering
	\caption{Owned versus shared object decomposition enabling scalable coordination through parallel execution.}
	\label{tab:object_decomposition}
	\begin{tabular}{|p{3.3cm}|p{2.5cm}|p{2.5cm}|p{2.5cm}|p{2.5cm}|}
		\toprule
		Object Type & State & Access & Update Frequency & Cost \\
		\midrule
		ContributorRegistry (owned) & Reputation, stake, privacy budget & Single owner write & Per-round per contributor & O(1) signature verification \\
		\hline
		RoundRegistry (shared) & Round config, committee & Multi-party consensus & Every 2 hours & O(M) BFT messages \\
		\hline
		ModelRegistry (shared) & Model lineage & Multi-party consensus & Per successful round & O(M) BFT messages \\
		\hline
		PolicyOracle (shared) & System rules & Multi-party consensus & Weekly to monthly & O(M) BFT messages \\
		\bottomrule
	\end{tabular}
\end{table}

\subsection{Round Lifecycle and Safety-Economics Coupling}

\begin{figure}[ht]
	\centering
	\includegraphics[width=0.85\textwidth]{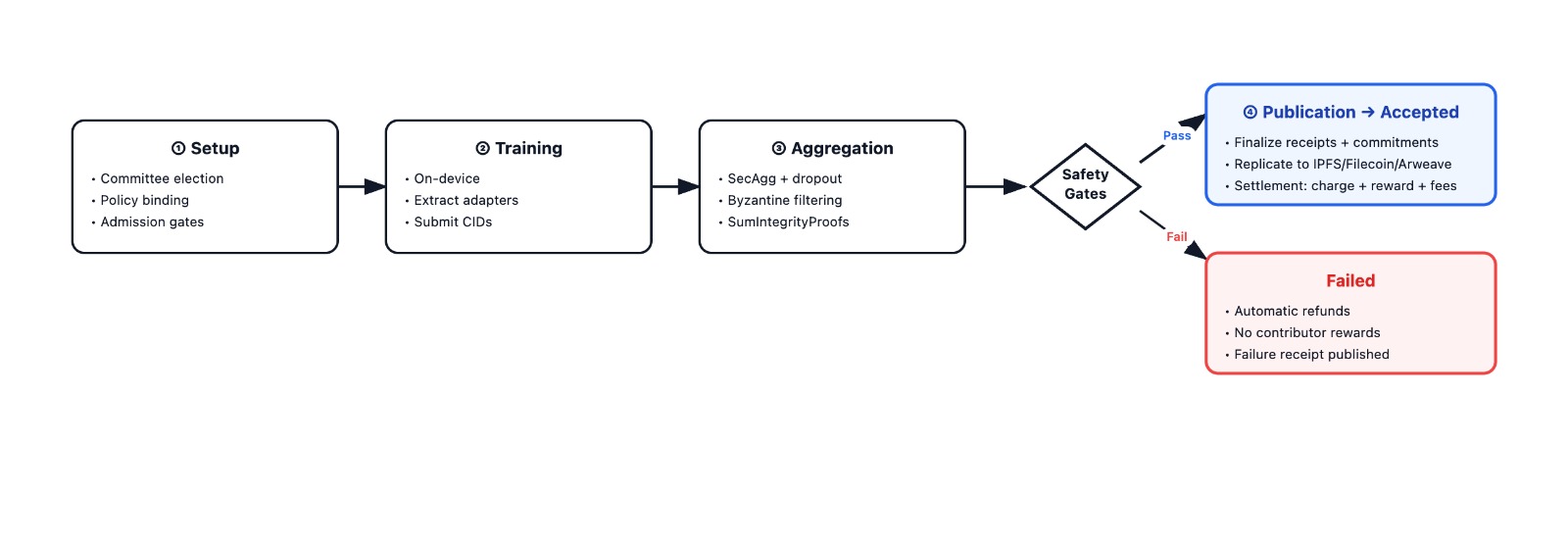}
	\caption{Round lifecycle state machine with deterministic phase transitions.}
	\label{fig:round_state_machine}
\end{figure}

Training proceeds through a deterministic state machine (Figure~\ref{fig:round_state_machine}). \textbf{Setup} (5 minutes): committee election, policy binding, contributor admission based on stake and attestation. \textbf{Training} (90 minutes): contributors train locally on private data that never leaves their devices, computing model improvements entirely on-device. Upon completion, contributors extract only low-rank adapter parameters from the final 1-2 transformer blocks, discarding all gradients and intermediate activations. \textbf{Crucially, contributors never transmit full model updates, raw gradients, or any information about their local training data}—only compressed adapter parameters (1-5 MB vs 80+ MB for full gradients), reducing communication 10-50× while fundamentally limiting gradient inversion attack surface \citep{Zhu2019}. The global model trunk remains frozen during local training, with improvements captured through adapter composition.

\textbf{Aggregation} (15 minutes): committee nodes combine masked adapter parameters via secure aggregation, reconstruct dropout masks for disconnected devices, apply Byzantine-robust filtering when variance exceeds policy thresholds, generate SumIntegrityProofs binding the aggregate to cryptographic commitments, and evaluate candidate models using pre-trained safety proxies. \textbf{Publication} (5 minutes): round receipts finalize with cryptographic commitments to all artifacts, models and proofs replicate to multi-provider storage (IPFS, Filecoin, Arweave), and economic settlement computes contributor rewards, committee fees, and any slashing penalties.

Terminal states: \textbf{Accepted} (all safety proxies pass thresholds, receivers charged, contributors rewarded proportionally) or \textbf{Failed} (any proxy exceeds threshold, automatic refunds to all receivers, no contributor rewards, complete failure receipt published documenting which proxies failed and by how much). Total round time of 2 hours meets service level objectives while preserving <10ms inference latency because training occurs asynchronously—users query on-device models with locally cached parameters, experiencing no latency from the training pipeline.

Safety-economics coupling eliminates trust in quality assessment. During Aggregation, committee nodes evaluate candidate models using pre-trained safety proxies (toxicity, PII leakage, fairness) specified in the active PolicyOracle bundle. If all proxy deltas remain below thresholds, the round transitions to Accepted and payments proceed. If any proxy exceeds thresholds, the round transitions to Failed, triggering automatic refunds. This makes dishonest evaluation economically irrational: operators passing failing models forfeit receiver payments while incurring committee operation costs.

\section{PROOF-CARRYING AGGREGATION}

Existing federated learning systems provide either confidentiality without integrity (SecAgg) or integrity without practical composition (zkFL for unweighted sums only). Elegant production systems require cryptographic receipts proving that weighted aggregation, dropout recovery, and Byzantine filtering were correctly applied, all within a single verifiable artifact.

We introduce \textbf{SumIntegrityProofs} composing these capabilities into unified cryptographic receipts. Each committee node \(j\) fetches masked adapter parameters from decentralized storage using CIDs submitted by admitted contributors to RoundRegistry and computes local masked sum \(S^{(j)} = \sum_{i \in admitted} w_i v_i^{(j)}\) where \(w_i\) are contributor-specific weights and \(v_i^{(j)}\) are masked adapter parameters. The node commits using vector Pedersen commitments (Pedersen, 1991): \( Com(S^{(j)}, r_S) \). Byzantine consensus combines these homomorphically: \( Com(S) = \prod_j Com(S^{(j)}) \), producing a global commitment to the masked aggregate.

\textbf{Why homomorphic commitments enable verifiable aggregation.} The homomorphic property enables verification without decryption: because \( Com(a + b) = Com(a) \cdot Com(b) \) for Pedersen commitments, auditors can verify that the committed aggregate equals the sum of committed node locals by checking the multiplicative relationship \( Com(S) = \prod_j Com(S^{(j)}) \), without ever learning individual values. This mathematical structure allows proving correct summation while preserving the confidentiality guarantee that SecAgg provides.

\textbf{Dropout and Byzantine handling.} When contributors drop out mid-round, the protocol reconstructs missing masks through Shamir secret sharing \citep{Bonawitz2017}, with reconstructed-set commitments binding which dropouts were recovered and preventing selective reconstruction. If aggregate variance exceeds policy thresholds indicating Byzantine behavior, the system applies coordinate-wise trimmed mean estimation \citep{Yin2018}, with robust method selection bound to PolicyOracle configuration locked before round execution. Both dropout recovery and Byzantine fallback integrate into the proof: auditors verify that declared policies were applied consistently rather than selectively manipulated.

\textbf{Dual proof path.} The homomorphic commitment path (default) provides fast verification with 10-100 KB proofs suitable for routine rounds but reveals aggregate structure (e.g., magnitude through commitment size). The zk-SNARK path \citep{BenSasson2014} for high-stakes rounds hides even aggregate structure through zero-knowledge properties, necessary when Byzantine nodes might exploit structural information to bias future rounds or when regulatory requirements mandate complete opacity.

\textbf{Verification and security.} Auditors verify receipts by: (1) fetching node-signed local commitments from content-addressed storage, (2) verifying signatures and commitment validity, (3) checking homomorphic combination \( Com(S) = \prod_j Com(S^{(j)}) \), (4) validating reconstructed-set commitments match dropout policy, (5) confirming Byzantine method selection matches PolicyOracle. Verification completes in under 10 seconds for homomorphic proofs. Security reduces to SecAgg's information-theoretic confidentiality under \(t-1\) collusion and Pedersen commitment computational binding under discrete logarithm hardness. The proof covers the entire aggregation pipeline—weighted combination, dropout recovery, Byzantine filtering, differential privacy application—in a single artifact, enabling any auditor to verify correctness without trust in the aggregator or access to individual contributions.

\section{GEOMETRIC NOVELTY DECOMPOSITION}

Sustainable federated learning requires fair compensation, but secure aggregation hides individual contributions, making payment mechanisms vulnerable to gaming. The ACE attack demonstrates that utility-based scoring—paying based on accuracy improvements—is fundamentally manipulable \citep{Xu2024}. Worse are replay attacks (resubmitting identical valuable updates across rounds for repeated payment) and sybil attacks (splitting updates across fake identities to claim multiple rewards), which existing systems cannot detect without storing all historical updates.

\begin{figure}[ht]
	\centering
	\includegraphics[width=0.85\textwidth]{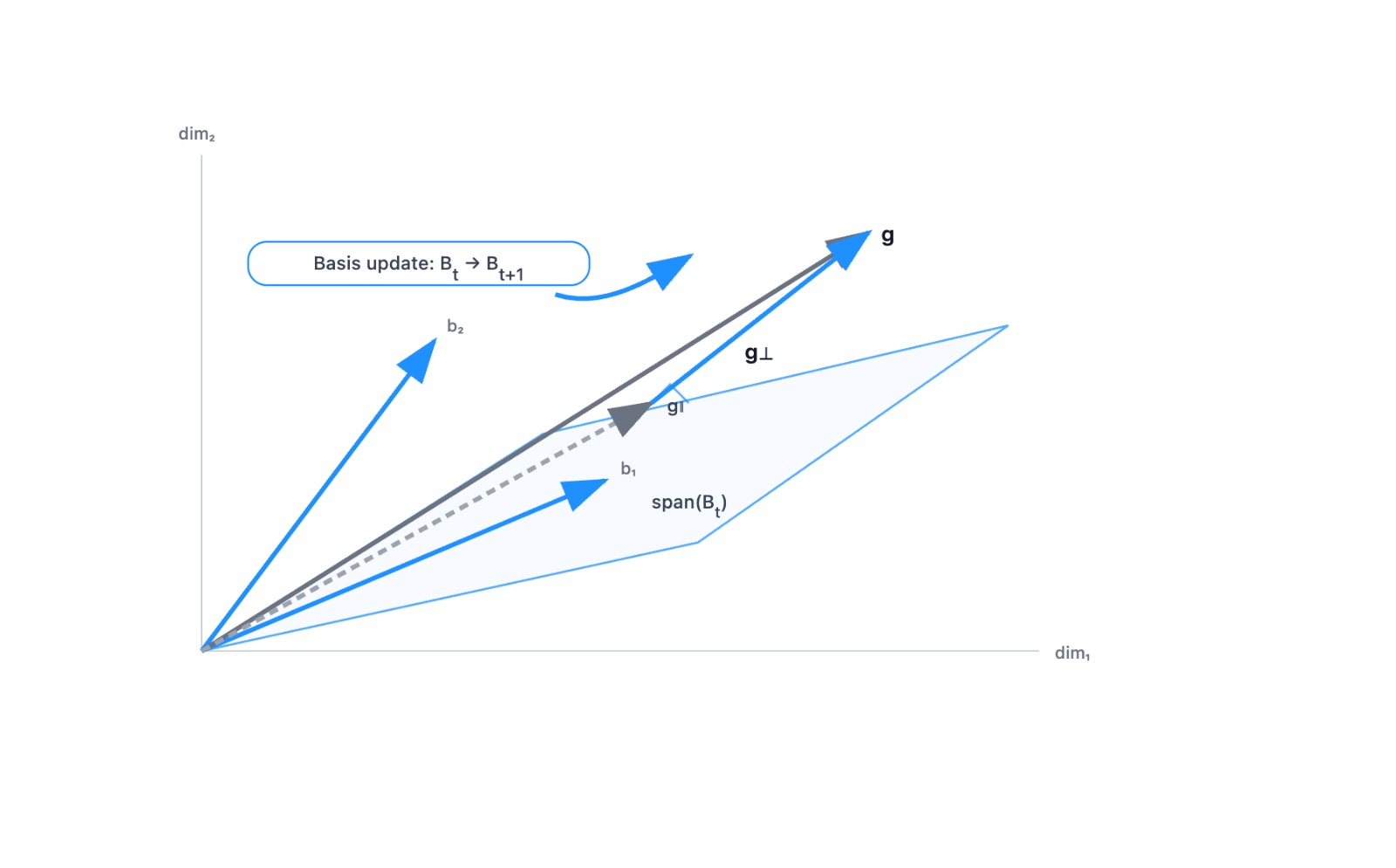}
	\caption{Geometric basis projection decomposing aggregate update $\mathbf{g}$ into parallel component $\mathbf{g}_{\parallel}$ (refinement of known patterns) and perpendicular component $\mathbf{g}_{\perp}$ (genuine exploration).}
	\label{fig:geometric_basis_projection}
\end{figure}

We introduce \textbf{geometric novelty decomposition} measuring directional contributions through basis projection. The system maintains a novelty basis \(B_t \in \mathbb{R}^{d \times k}\) representing already-explored directions in the model's parameter space, where \(d\) is the adapter dimension and \(k\) is the basis size. When a new aggregate update \(g\) arrives, it decomposes into orthogonal components (Figure~\ref{fig:geometric_basis_projection}):

\[
g = g_\parallel + g_\perp
\]

where \(g_\parallel = B_t(B_t^T g)\) represents refinement of known patterns (parallel component) and \(g_\perp = g - g_\parallel\) represents genuine exploration (perpendicular component). The novelty score measures:

\[
\phi_t = \frac{\| g_\perp \|}{\| g \| + \epsilon}
\]

where \(\epsilon\) prevents division by zero. The system rewards only the perpendicular component proportional to \(\| g_\perp \|\), then rotates the basis to incorporate this new direction: \(B_{t+1} \leftarrow update(B_t, g_\perp / \| g_\perp \|)\), making that direction permanently "known" for future rounds. Temporal smoothing \(\tilde{\phi}_t = \lambda \tilde{\phi}_{t-1} + (1-\lambda) \phi_t\) with \(\lambda = 0.7\) reduces noise from round-to-round variance.

\textbf{Attack resistance through geometric structure.} Replay attacks fail because a previously rewarded update now lies entirely in the known subspace \(B_{t+1}\), producing \(g_\perp = 0\) and \(\phi_t = 0\), yielding zero reward. An adversary resubmitting a valuable update from round 10 in rounds 11, 12, and 13 collects payment only once—the first time, before that direction enters the basis. Sybil attacks fail because novelty is measured on the aggregate: splitting a perpendicular update across 100 fake identities produces 100 individual updates that, when summed during aggregation, yield the identical aggregate \(g\) and therefore identical \(g_\perp\) and \(\phi_t\). The total reward pool remains unchanged, now divided by 100 instead of claimed once. Since identity creation costs stake deposits (minimum \$10-50 per identity), splitting becomes economically irrational.

\textbf{Game-theoretic equilibrium.} Honest contributors maximize rewards by genuinely exploring parameter space rather than gaming mechanisms. An honest contributor earning reward \(r_{honest} = r_{base} + r_{novelty} \cdot \| g_\perp \|\) dominates an adversary attempting replay (earning only \(r_{base}\) after first submission) or sybil splitting (earning \(r_{honest}/n\) for \(n\) identities while paying \(n \times stake costs\)). The mechanism preserves privacy because novelty is computed on aggregates, never exposing individual contributions. The geometric approach trades semantic guarantees for computational efficiency: an update perpendicular to \(B_t\) is geometrically novel but may be semantically unhelpful (e.g., adversarial perturbations). Safety gates (Section 8) provide backstop protection by rejecting rounds where models degrade on held-out test sets.

\section{TIME-LOCKED GOVERNANCE}

    Traditional federated learning enables retroactive rule manipulation: aggregators adjust safety thresholds, privacy budgets, or admission criteria after observing training outcomes. Simply recording rules on-chain proves insufficient if rules can change arbitrarily—an aggregator could propose a new policy in round 100, activate it immediately, and apply it retroactively to rounds 95-99.

We introduce \textbf{PolicyOracles} with activation snapshots that make retroactive manipulation cryptographically provable fraud. Every policy proposal undergoes mandatory lock periods: \(T_{lock} \geq 5\) rounds minimum between proposal and activation. A policy proposed in round 95 cannot activate before round 100. When round 100 executes, its receipt cryptographically binds to the policy version active at round 100 through content-addressed references (IPFS CIDs) stored immutably on-chain. Attempting to change round 100's governing policy requires rewriting blockchain history, which Byzantine consensus prevents \citep{Castro1999}.

\textbf{Policy bundles} compose five components: safety thresholds (toxicity, PII, fairness limits), differential privacy parameters (\(\epsilon\) per round, \(\delta\) global, clipping bounds), admission criteria (minimum stake, attestation requirements, deadlines), aggregation rules (robust method selection, quantization), and novelty economics (basis rotation schedule, reward splits). Each bundle receives a content identifier binding all parameters atomically—changing any single threshold requires proposing an entirely new bundle subject to \(T_{lock}\).

\textbf{Governance participation and stakeholder classes.} \textbf{PolicyOracle} governance operates through multi-stakeholder voting where human participants—not nodes—propose and vote on policies. Stakeholder classes include: (1) Contributors with weight proportional to reputation earned through geometric novelty, (2) Receivers with weight proportional to payment volume, (3) Validator operators with weight proportional to stake, and (4) External stakeholders participating through token-weighted mechanisms. Critically, nodes themselves do not vote—validators are technical infrastructure operated by humans, and governance power derives from human stake, not node count. This prevents Sybil governance attacks where one operator running multiple nodes cannot claim multiple votes. Proposals require minimum stake thresholds, undergo public voting periods, and face mandatory time-lock periods before activation. Vote weights use quadratic scaling to prevent plutocracy while maintaining stake alignment.

\textbf{Progressive decentralization} adapts governance as the system matures. Phase 0 (rounds 0-500): multisignature control with 3-of-5 operators, \(T_{lock} = 5\) rounds, rapid iteration for bug fixes. Phase 1 (rounds 500-2000): parameter council of 9 elected contributors controls economic parameters (reward splits, novelty \(\beta\)), while multisig retains safety-critical authority, \(T_{lock} = 10\) rounds for safety bundles. Phase 2 (rounds 2000+): DAO with stake-weighted voting, quorum = 20\% for safety changes, \(T_{lock} = 20\) rounds, supermajority 66\% for constitutional changes.

\textbf{Emergency controls} allow temporary admission or publication halts with automatic expiry (maximum 72 hours), logged justifications, and concurrent halt limits (1 per cohort + 1 global). Halts transitioning to \textbf{Failed(AutoExpired)} trigger automatic refunds, preventing indefinite operational freezes while maintaining emergency response capability.

For full schema definitions and encoding constants enabling bit-exact, cross-implementation verification of policy bundles and receipts, see Appendix A (Core Data Schemas).

    \section{PRIVACY AND SAFETY COMPOSITION}

The architecture composes four privacy layers operating on different information planes. 
\textbf{Adapter-only transmission} reduces dimensionality 10--50$\times$, limiting gradient inversion attack surface \citep{Zhu2019}. 
\textbf{Secure aggregation} ensures committee members observe only masked sums through information-theoretic confidentiality under $t-1$ collusion \citep{Bonawitz2017}. 
\textbf{Differential privacy} bounds statistical inference through gradient clipping ($C = 1.0$) and calibrated Gaussian noise ($\sigma = 0.5$), with R{\'e}nyi accounting tracking cumulative privacy loss ($\varepsilon = 1.0$ per round, $\delta = 10^{-6}$) \citep{Mironov2017}. 
Differential privacy provides a dual guarantee: privacy preservation and algorithmic stability \citep{Wang2016}. 
Models trained under $(\varepsilon,\delta)$-DP constraints generalize well even without explicit regularization, as the privacy mechanism bounds the influence of any single contributor. 
\cite{Chaudhuri2011} show that objective perturbation---applying noise during aggregation rather than to final models---achieves superior utility--privacy tradeoffs for convex objectives, with noise scaling as $O(1/\sqrt{n})$ for $n$ contributors. 
Our choice to apply differential privacy during the aggregation phase follows this principle. 
\textbf{Byzantine-robust aggregation} applies coordinate-wise trimmed mean when variance exceeds $\theta = 90^{th}$ percentile, limiting adversarial influence to bounded fractions \citep{Yin2018}.

\textbf{Safety gates} couple model quality to economics. Committee nodes evaluate candidate models using pre-trained proxies assessing toxicity, PII leakage, and fairness on held-out test sets. If all proxy deltas remain below PolicyOracle thresholds, the round transitions to Accepted (payments proceed). If any proxy exceeds thresholds, the round transitions to Failed (automatic refunds, no contributor rewards, complete failure receipt published). This makes dishonest evaluation economically irrational: operators passing failing models forfeit receiver payments while incurring operation costs. Composition risks persist: cross-layer information leakage (correlating DP noise with Byzantine filtering decisions) and side-channel vulnerabilities (timing, traffic patterns) require deployment-level mitigations including submission jitter and traffic padding.

\section{INCENTIVE ECONOMY}

\begin{figure}[ht]
	\centering
	\includegraphics[width=0.85\textwidth]{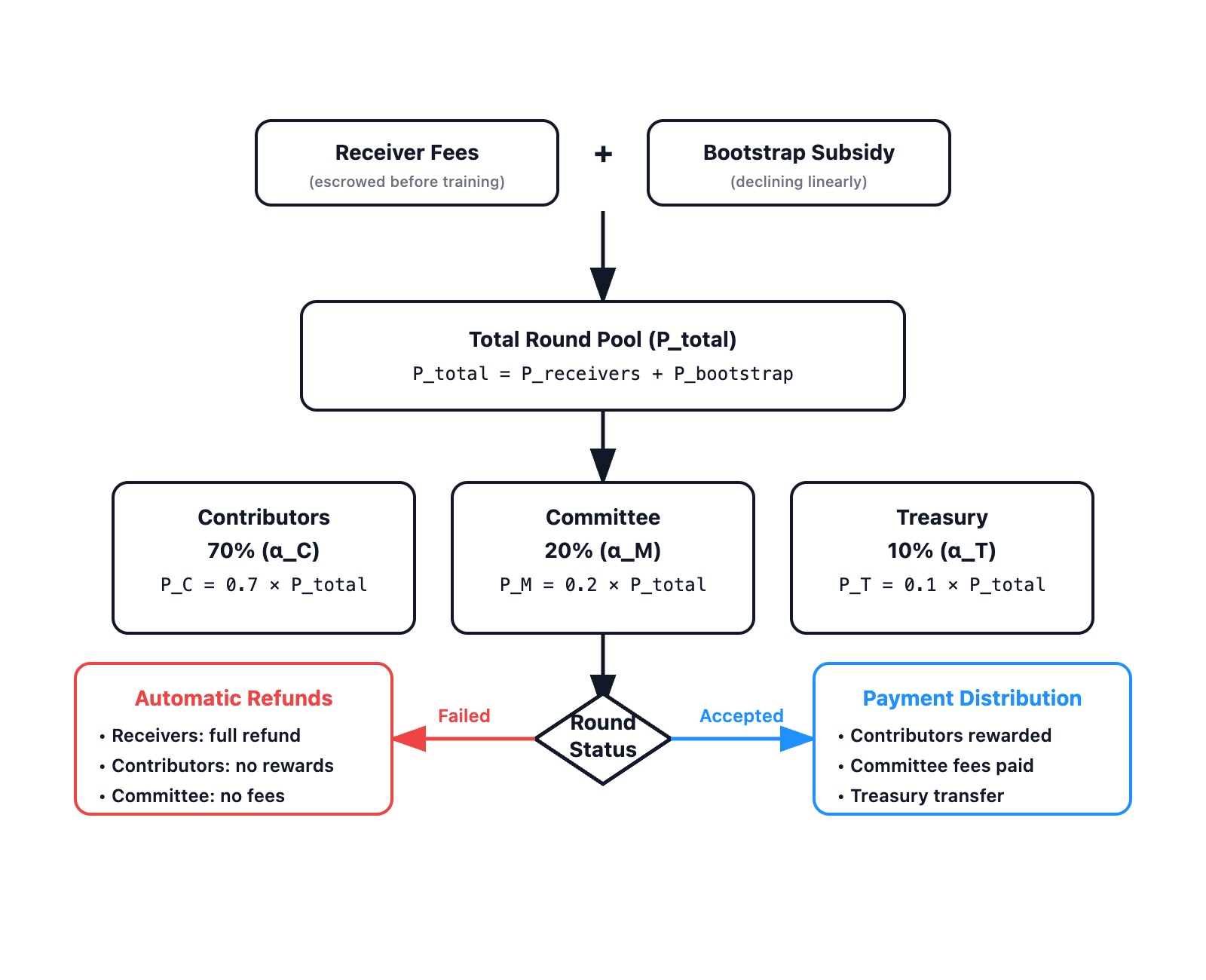}
	\caption{Incentive economy flow showing safety--economics coupling. Round pools form from escrowed receiver fees and bootstrap subsidies, split across contributors (70\%), committee (20\%), and treasury (10\%). Failed rounds trigger automatic refunds; accepted rounds distribute payments proportionally.}
	\label{fig:incentive_flow}
\end{figure}

Round pools form from escrowed receiver fees and bootstrap subsidies (Figure~\ref{fig:incentive_flow}): \(P_{total} = P_{receivers} + P_{bootstrap}\), where receivers lock micro-fees (\$0.001-\$0.01 per round) into AccessLedgerEscrow before training, binding payment to specific policy versions. The pool splits across three participant classes: contributors (\(\alpha_C = 70\%\)), committee (\(\alpha_M = 20\%\)), treasury (\(\alpha_T = 10\%\)).

Individual contributor rewards combine three components: base participation reward (\(r_{base}\) per admitted contributor), reputation-weighted quality adjustments (multiplier \(\rho \in [0.8, 1.2]\) based on historical contribution quality), and novelty-indexed bonuses (\(\beta \cdot P_C \cdot \tilde{\phi}_t\) distributed proportionally): 

\[
r_i = r_{base} \cdot \rho + r_{quality,i} + r_{nov,i}
\]

Reputation multipliers increase by 0.05 per successful round (capped at 1.2) and decay by 0.1 for rounds where contributions exhibit high variance relative to the aggregate. Committee members receive uniform fees (\(P_M / M\) for \(M\) nodes) only for \textbf{Accepted} rounds, aligning infrastructure incentives with model quality.

\textbf{Failed} rounds trigger three automatic consequences: escrowed receiver fees refund immediately through smart contract execution, contributors receive zero rewards because no value was delivered, and complete failure receipts document which safety proxies failed and by how much. This safety-economics coupling makes dishonest evaluation economically irrational.

Slashing mechanisms penalize provable faults with proportional stake reductions: invalid \textbf{SumIntegrityProofs} (30\% stake), selective mask reconstruction favoring specific contributors (20\% stake), and liveness failures preventing round completion (10\% stake). Merkle-rooted payout trees enable deterministic verification: auditors sample \(k \geq 100\) contributor rewards, recompute \(r_i\) from receipt data, validate Merkle inclusion proofs, and extrapolate to verify total conservation \(P_{total} = P_C + P_M + P_T + dust\), with all residuals explicitly accounted in dust fields using 16-bit fixed-point arithmetic with ties-to-zero rounding.

    \section{SECURITY ANALYSIS}
\label{sec:security_analysis}

\subsection{Compositional Correctness}

The architecture's security reduces to four established primitives through compositional arguments demonstrating that mechanisms reinforce rather than undermine each other.

\textbf{Theorem 1 (Privacy preservation).} The system preserves \((\epsilon, \delta)\)-differential privacy and information-theoretic confidentiality of individual contributions under the assumptions that: (1) at most \(t-1\) committee members collude where \(t = \lceil M/2 \rceil\), (2) \textbf{SecAgg} mask reconstruction follows Bonawitz et al. (2017), (3) differential privacy noise follows Rényi accounting \citep{Mironov2017} with declared parameters, and (4) \textbf{adapter-only transmission} limits dimensionality to \(d_{\text{adapter}} \ll d_{\text{full}}\).

\textit{Proof sketch.} Confidentiality reduces to \textbf{SecAgg}'s information-theoretic security: under \(t-1\) collusion, colluding nodes observe masked sums \(S^{(j)}\) but cannot reconstruct individual contributions \(v_i\) because masks cancel only in aggregate. Differential privacy composition follows from per-round clipping (\(C=1.0\)) and Gaussian noise (\(\sigma=0.5\)) with Rényi accountant tracking cumulative \(\epsilon\). \textbf{Adapter-only transmission} reduces gradient inversion attack surface by limiting dimensionality: reconstruction attacks require solving underdetermined systems with \(d_{\text{adapter}}\) unknowns (1-5M parameters) versus \(d_{\text{full}}\) (80M+ parameters), exponentially reducing attack success probability \citep{Zhu2019}. The composition holds because \textbf{SecAgg} operates on adapter parameters after DP noise application, with privacy budgets tracked independently per contributor through ContributorRegistry objects. \(\square\) The storage architecture preserves privacy by requiring committee nodes to fetch adapters from public IPFS using CIDs, where adapters remain masked throughout transmission and storage. Storage providers observe only encrypted blobs without CIDs-to-contributor mappings, preventing linkage attacks. The content-addressed nature of IPFS ensures integrity: any modification to stored adapters would result in a different CID, immediately detectable during verification.

\textbf{Theorem 2 (Integrity verification).} For any round receipt \(R\) with \textbf{SumIntegrityProof} \(\pi\), an auditor can verify with probability \(1 - 2^{-\lambda}\) for security parameter \(\lambda\) that the aggregate \(S\) committed in \(R\) equals \(\sum_j w_j S^{(j)}\) where \(S^{(j)}\) are node-signed local sums and \(w_j \in [w_{\text{min}}, w_{\text{max}}]\) are policy-bounded weights, assuming: (1) Pedersen commitment binding under discrete logarithm hardness, (2) Byzantine consensus with \(f < M/3\) faulty nodes, and (3) node signatures are unforgeable under chosen-message attacks.

\textit{Proof sketch.} Verification checks \( \text{Com}(S) = \prod_j \text{Com}(S^{(j)}) \). By Pedersen commitment binding, finding \(S' \neq S\) with \( \text{Com}(S') = \text{Com}(S) \) requires solving the discrete logarithm problem, computationally infeasible under standard assumptions. Byzantine consensus ensures at least \(2f + 1\) honest nodes sign consistent local commitments \(\text{Com}(S^{(j)})\). Weight policy binding ensures \(w_j \in [w_{\text{min}}, w_{\text{max}}]\) through Merkle inclusion proofs against \( \text{weights}_{\text{root}} \). Dropout reconstruction commitments bind reconstructed set \(R_{\text{drop}}\), preventing selective reconstruction: changing \(R_{\text{drop}}\) requires forging new Merkle proofs, detectable through root mismatch. Byzantine-robust fallback selection binds to PolicyOracle configuration through \(\text{policy}_{\text{cid}}\) reference, making ex post facto robust method changes require blockchain history rewriting. \(\square\)

\subsection{Incentive Robustness}

The geometric novelty mechanism resists replay and sybil attacks through mathematical properties of basis projection rather than cryptographic assumptions.

\textbf{Theorem 3 (Replay resistance).} An adversary submitting aggregate update \(g\) in round \(t\) and receiving novelty reward \(\| g_\perp \|\) cannot profitably resubmit \(g\) in round \(t' > t\), because \(g \in \text{span}(B_{t'})\) after basis rotation, yielding \(\| g_\perp \| = 0\) and zero novelty reward.

\textbf{Proof sketch.} After round \(t\), the basis rotates: \(B_{t+1} \leftarrow [B_t \ | \ g_\perp/\| g_\perp \|]\) with the oldest direction dropped. Therefore, \(g_\perp \in \text{span}(B_{t+1})\). In round \(t'\), the decomposition \(g = g_\parallel' + g_\perp'\) with \(g_\parallel' = B_{t'}(B_{t'}^T g)\) satisfies \(g_\perp' = 0\) because \(g \in \text{span}(B_{t'}) \subseteq \text{span}(B_{t+1})\). Novelty score \(\phi_{t'} = \| g_\perp' \| / (\| g \| + \epsilon) = 0\), yielding zero novelty bonus. The adversary pays computational cost \(C_{\text{train}}\) for local training and stake deposit \(S_{\text{min}}\) but receives only \(r_{\text{base}}\) without novelty component, making replay unprofitable when \(r_{\text{base}} < C_{\text{train}} + S_{\text{min}} \cdot \text{interest rate}\). \(\square\)

\textbf{Theorem 4 (Sybil resistance).} An adversary splitting aggregate update \(g\) across \(n\) fake identities \(\{g_1, \ldots, g_n\}\) where \(\sum_i g_i = g\) receives total novelty reward equal to honestly submitting \(g\) from a single identity, while paying \(n \times\) identity creation costs.

\textbf{Proof sketch.} Novelty is measured on the aggregate after secure aggregation: \(g_{\text{agg}} = \sum_i g_i = g\). Decomposition \(g_{\text{agg}} = g_\parallel + g_\perp\) yields identical perpendicular component regardless of how individual contributions partition. Total novelty pool \(\beta \cdot P_C \cdot \tilde{\phi}_t\) distributes proportionally to \(\| g_\perp \|\), which remains unchanged. The adversary receives total reward \(R_{\text{total}} = n \cdot r_{\text{base}} + \beta \cdot P_C \cdot \tilde{\phi}_t \cdot (\| g_\perp \| / \| g_{\text{agg}} \|)\) while paying \(n \cdot S_{\text{min}}\) in stake deposits and \(n \cdot C_{\text{attestation}}\) in attestation costs. Honest submission yields \(R_{\text{honest}} = r_{\text{base}} + \beta \cdot P_C \cdot \tilde{\phi}_t \cdot (\| g_\perp \| / \| g \|)\). Sybil attack is profitable only when \(R_{\text{total}} - n \cdot (S_{\text{min}} + C_{\text{attestation}}) > R_{\text{honest}}\), which reduces to \((n-1) \cdot r_{\text{base}} > n \cdot (S_{\text{min}} + C_{\text{attestation}})\). For \(S_{\text{min}} \geq r_{\text{base}}\) and \(C_{\text{attestation}} > 0\), this inequality never holds, making sybil splitting economically irrational. \(\square\)

\subsection{Governance Security}

\textbf{Property 1 (Non-retroactivity).} For any round \(r\) with receipt binding to \(\text{policy\_cid}_r\) active at round \(r\), attempting to change the governing policy requires either: (1) rewriting blockchain history to alter the on-chain \(\text{policy\_cid}_r\) reference (prevented by Byzantine consensus finality), or (2) altering the content-addressed policy bundle at \(\text{policy\_cid}_r\) (prevented by cryptographic hash collision resistance).

\textbf{Property 2 (Lock period enforcement).} PolicyOracle smart contracts enforce \(T_{\text{lock}} \geq 5\) rounds between \(\text{propose\_round}\) and \(\text{activation\_round}\) through on-chain verification: activation transactions failing the check \(\text{activation\_round} \geq propose\_round + T_{lock}\) revert automatically. This provides contributors with at least 5 rounds (10 hours at 2-hour cadence) to review proposed changes, assess implications, and exit if policies become unacceptable, preventing surprise rule changes.

\subsection{Trust Assumptions and Limitations}

The security analysis assumes: (1) Byzantine fault tolerance with \(f < M/3\) faulty committee members, where illustrative \(M=7\) supports \(f=2\); production deployments should expand to \(M=20-100\) nodes providing stronger collusion resistance; (2) cryptographic primitives including SHA-256 collision resistance and discrete logarithm hardness remain computationally infeasible; (3) TEE-based attestation for client-side safety filtering inherits commodity TEE trust assumptions including microcode supply chain integrity; (4) side-channel attacks including timing analysis and traffic pattern observation lie outside the threat model, requiring deployment-specific mitigations. The architecture provides defense-in-depth where compromising any single mechanism reveals limited information and cannot forge system-wide state, but does not claim absolute security against all adversaries. Adaptive machine learning adversaries iteratively crafting updates that pass safety gates yet subtly degrade quality receive partial mitigation through ensemble proxy evaluation but require ongoing red-teaming.

Formalized statements of the core guarantees and abbreviated proof sketches are provided in Appendix B.

\section{COST AND PERFORMANCE ANALYSIS}
\label{sec:cost_performance_analysis}

\subsection{Computational Costs}

\begin{table}[ht]
	\centering
	\caption{Decomposed Per-Round Computational Costs at 10,000 Contributors with 20M Adapter Parameters}
	\label{tab:computational_costs}
	\begin{tabular}{|p{2.5cm}|p{2.5cm}|p{2.5cm}|p{2.5cm}|p{2.5cm}|}
		\toprule
		Component & Operation & Time & Hardware & Notes \\
		\midrule
		Contributor & Local training (10 epochs) & 3-8 min & On-device GPU/NPU & Varies by device tier \\
		\hline
		& Adapter extraction & 2-5 sec & CPU & Low-rank projection \\
		\hline
		& SecAgg masking & 0.3-0.8 sec & CPU & Pairwise seed generation \\
		\hline
		& Upload (1-5 MB) & 10-50 sec & 1-10 Mbps uplink & Network-bound \\
		\hline
		Committee (per node) & Receive masked updates & 5-15 sec & Network I/O & 10K x 1-5 MB \\
		\hline
		& Local sum computation & 3-8 sec & CPU & Weighted aggregation \\
		\hline
		& Pedersen commitment & 5-12 sec & CPU & Vector commitment \\
		\hline
		& Consensus (BFT) & 10-30 sec & Network & M=7 nodes, 3 rounds \\
		\hline
		& Dropout reconstruction & 15-45 sec & CPU & Shamir sharing, t-1 dropouts \\
		\hline
		& Byzantine filtering & 8-20 sec & CPU & Coordinate-wise trimmed mean \\
		\hline
		& Proof generation & 8-15 sec & CPU & Homomorphic path \\
		\hline
		Auditor & Proof verification & 6-12 sec & CPU & Homomorphic commitment check \\
		\hline
		& Safety re-evaluation & 2-4 min & GPU & Optional, full proxy rerun \\

		\bottomrule
	\end{tabular}
\end{table}

\noindent
\textbf{Table~\ref{tab:computational_costs}:} Per-round computational costs showing committee aggregation and safety evaluation dominate latency (15 minutes total), while contributor training occurs asynchronously without blocking inference.

\textbf{Critical path latency.} Setup (5 min) + Training (90 min, asynchronous) + Aggregation (15 min) + Publication (5 min) = 2 hours total round time. Training dominates wall-clock time but occurs on-device without coordinator involvement. Aggregation dominates critical path latency (15 minutes) due to safety proxy evaluation requiring GPU inference across 150-1500 prompts (3 proxies × 50-500 prompts). Proof generation (8-15 sec) and verification (6-12 sec) add negligible overhead compared to safety evaluation.

\subsection{Economic Sustainability}

\textbf{On-chain costs.} Owned-object updates (ContributorRegistry modifications) cost O(1) per contributor: signature verification (0.5ms) + state write (2ms) \(\approx\) 0.0025 gas units per update at fixed pricing. For 10,000 contributors, parallel owned-object updates cost 10,000 × 0.0025 = 25 gas units. Shared-object updates (RoundRegistry phase transitions, ModelRegistry lineage append, PolicyOracle activation) cost O(M) for M=7 committee nodes: Byzantine consensus (3 rounds × 7 nodes × 21 messages = 147 BFT messages) + state writes \(\approx\) 0.5 gas units per shared update. Per round: 25 gas units (owned) + 2 gas units (4 shared updates × 0.5) = 27 gas units \(\approx\) \$0.027 at fixed pricing of \$0.001 per gas unit, yielding \$0.0000027 per contributor per round.

\textbf{Off-chain storage.} Contributors upload masked adapters (1-5 MB each) to IPFS, with total per-round storage of 10,000 contributors × 2 MB average = 20 GB. Committee publishes aggregated models (80 MB) and receipts (100 KB). IPFS pinning (fast retrieval) costs \$0.01-0.05 per GB-month. Filecoin deals (long-term provable storage) cost \$0.05-0.15 per GB-month. Arweave permanent storage (receipts, final models) costs \$0.50-2.00 per GB one-time. Per round: 20 GB (adapters, temporary) + 80 MB (model) + 100 KB (receipt) + 50 KB (proof). Contributor adapters can be pruned after aggregation completes, reducing long-term storage to aggregate models only (80 MB per round). Monthly cost (30 rounds): 2.4 GB aggregate models × (\$0.05 IPFS + \$0.10 Filecoin) = \$0.36/month for recurring storage, plus Arweave archival \$2.04/month (85 MB × 30 rounds × \$1/GB one-time amortized).

\textbf{Total per-round cost.} On-chain (27 gas units \(\sim\) \$0.027) + storage (\$0.098) + committee operation (\$2-5 amortized across validator set) \(\sim\) \$2.125-5.125 per round for 10,000 contributors = \$0.0002-0.0005 per contributor. Receiver micro-fees of \$0.001-0.01 per round provide 2-50× cost recovery, enabling economic sustainability even during bootstrap phases with subsidies. Bootstrap subsidy of \$50/round (declining linearly over 1000 rounds) bridges early-phase adoption, after which receiver fees alone sustain operations.

\subsection{Scalability Analysis}

\textbf{Object-centric coordination scaling.} Owned-object updates execute in parallel with O(N) total cost for N contributors. Shared-object updates cost O(M) for M committee nodes but occur infrequently (4 updates per 2-hour round). Total coordination cost: O(N) + O(M) \(\approx\) O(N) since N >> M (10,000 contributors vs 7-100 committee nodes). Traditional blockchains serializing all updates exhibit O(N²) cost as each update waits for all prior updates to process through consensus.

\textbf{Committee aggregation scaling.} Aggregation computation is O(N · d) for N contributors with d-dimensional adapters: summing N vectors of dimension d = 1-5M. With d fixed, aggregation scales linearly in N. For M=7 committee nodes each processing N/M contributors, parallel aggregation across nodes maintains O(N/M · d) per-node cost. Byzantine consensus cost O(M²) for M nodes remains tractable for M=7-100. Safety proxy evaluation cost is O(P · E · d) for P proxies, E evaluation prompts, and d model parameters, independent of contributor count N—evaluating 3 proxies on 500 prompts costs the same whether 1,000 or 100,000 contributors submitted updates.

\textbf{Storage scaling.} Content-addressed artifacts scale horizontally: adding storage providers increases redundancy and retrieval bandwidth without protocol modifications. Multi-provider replication (minimum 3 providers) ensures availability despite single-provider failures. Per-round storage cost remains constant (85 MB × provider costs) independent of contributor count N, as only the aggregate model and proof require storage, not individual updates.

\textbf{Empirical validation gap.} The cost and performance analysis derives from analytical models informed by published benchmarks for DAG-based consensus systems and SecAgg implementations \citep{bonawitz2017practical}, using realistic parameters (10K contributors, 20M adapters, 7-node committees). However, empirical validation through testnet deployment with thousands of edge devices, sustained operation over hundreds of rounds under realistic churn, and adversarial stress testing remains critical future work. The analysis demonstrates architectural feasibility and acceptable cost bounds but cannot predict all real-world performance characteristics without implementation.

Operational parameters, cost assumptions, and scalability limits used throughout the evaluation are enumerated in Appendix C (Implementation Parameters)

\section{DISCUSSION}

\subsection{Design Trade-offs}

\textbf{Custom blockchain versus public mainnet.} The architecture adopts a purpose-built blockchain implementing DAG-based Byzantine consensus rather than deploying on existing public infrastructure.. This trade-off prioritizes cost predictability and FL-specific optimization at the expense of operational overhead. Public mainnets offer mature security (Ethereum: \$200B+ market cap securing consensus), established tooling (Remix, Hardhat, extensive libraries), and network effects (interoperability with DeFi, existing user bases). However, they impose fundamental mismatches: dynamic gas markets create 10-100× cost volatility during congestion (Ethereum gas: 20 gwei baseline, 500+ gwei during NFT mints), making \$0.001 consumer micro-fees infeasible when round costs fluctuate unpredictably. Governance misalignment means FL-specific protocol upgrades (extending block size for large receipts, tuning consensus timing for 2-hour rounds) compete with DeFi priorities controlling on-chain governance. Privacy exposure through permissionless validators increases metadata analysis risk compared to permissioned validator sets with explicit geographic and institutional diversity. The custom blockchain enables fixed gas pricing (\$0.0000027 per contributor) and governance aligned to FL requirements, but requires recruiting validators, coordinating security patches, and maintaining independent infrastructure. Production deployments must weigh predictable micro-fee economics against operational complexity.

\textbf{Geometric novelty versus Shapley values.} The architecture measures contribution through basis projection rather than Shapley value approximation. Shapley values provide theoretically optimal fair attribution by computing each contributor's marginal contribution across all possible coalitions, but require exponential evaluations (\(2^N\) subsets for \(N\) contributors). Monte Carlo approximations reduce this to polynomial complexity but remain impractical at 10,000+ scale (Ghorbani and Zou, 2019). Geometric novelty achieves \(O(d)\) projection per round for \(d\)-dimensional adapters, enabling real-time computation. However, the trade-off sacrifices semantic guarantees: an update perpendicular to basis \(B_t\) is geometrically novel but may be semantically unhelpful (adversarial perturbations orthogonal to known directions). Safety gates provide backstop protection by rejecting rounds where models degrade on held-out test sets, but sophisticated adversaries might craft updates that pass safety thresholds yet incrementally shift behavior over many rounds. The geometric approach prioritizes computational efficiency and replay/sybil resistance over semantic precision, accepting that empirical validation must demonstrate correlation between geometric novelty and genuine learning progress.

\textbf{Adapter-only versus full model transmission.} Restricting transmission to low-rank adapters reduces communication 10-50× and limits gradient inversion attack surface but constrains global model learning capacity. Full fine-tuning enables arbitrary representational changes across all model layers, while adapters bottleneck updates through low-rank subspaces (Houlsby et al., 2019; Hu et al., 2022). Mobile FL literature demonstrates adapters capture 80-90\% of full fine-tuning quality for common edge tasks (autocomplete, recommendations, content ranking), making this trade-off acceptable for many deployments. However, adapters may prove insufficient for tasks requiring fundamental architectural changes or learning entirely new capabilities. The architecture could support full gradient transmission through extended aggregation time budgets and increased bandwidth costs, but loses the privacy enhancement adapters provide. Production deployments must assess whether adapter expressiveness suffices for target tasks.

\subsection{Limitations and Future Work}

\textbf{Empirical validation gap.} This work demonstrates architectural feasibility through rigorous analysis reducing security to established primitives, analytical cost models deriving performance bounds, and worked examples with realistic parameters. However, empirical validation through testnet deployment with thousands of edge devices, adversarial red-teaming against incentive mechanisms, sustained operation over hundreds of rounds under realistic churn, and user studies on micro-fee acceptance remains critical future work. The 2024 BC-FL survey notes that fewer than five systems have progressed beyond proof-of-concept to production deployments serving 1000+ sustained participants (Ning et al., 2024), indicating empirical validation at scale represents a field-wide challenge rather than limitation specific to this work.

\textbf{Adaptive adversaries.} The threat model addresses static attacks (replay, sybil, Byzantine updates) but provides partial mitigation against adaptive adversaries who iteratively refine attacks based on system responses. Sophisticated adversaries might craft updates that simultaneously achieve high geometric novelty through basis orthogonality, pass safety gates through incremental degradation below detection thresholds, and survive Byzantine filtering through statistical similarity to honest updates, yet subtly bias model behavior over many rounds. Defending against such adaptive threats requires ongoing red-teaming, anomaly detection across round sequences, and potentially ML-based attack classification—active research areas beyond this work's scope.

\section{RELATED SYSTEMS}

Table~\ref{tab:comparison_systems} positions this work relative to representative BC-FL systems.

\begin{table}[ht]
	\centering
	\caption{Comparison with representative blockchain-federated learning systems, showing this work composes capabilities that existing systems provide in isolation.}
	\label{tab:comparison_systems}
	\begin{tabular}{|p{1.5cm}|p{3.0cm}|p{3.0cm}|p{3.0cm}|p{3.0cm}|}
		\toprule
		System & Verifiable Aggregation & Replay/Sybil Resistant & 10K+ Scaled Validated & Privacy Composition \\
		\midrule
		BlockFLA & Committee signatures & Utility-based (gameable) & Simulation only & Single-param time-locks \\
		\hline
		FLChain & Merkle commitments & Reputation with decay & 50 nodes (sim) & No time-locks \\
		\hline
		BLADE-FL & Committee attestation & Stake-weighted voting & Testnet (100 nodes) & Certificate Transparency adapted \\
		\hline
		This work & Proof-carrying (weighted + dropout + Byzantine) & Geometric novelty (basis projection) & Analytical (10K contributors) & PolicyOracle bundles with activation snapshots \\		
		\bottomrule
	\end{tabular}
\end{table}

BlockFLA (prior work) introduces adaptive contribution evaluation through gradient similarity but relies on utility-based scoring vulnerable to ACE attacks \citep{Xu2024}. FLChain demonstrates hierarchical sharding at 50-node scale in simulation but does not address replay/sybil resistance or provide cryptographic aggregation proofs. BLADE-FL proposes committee attestation for verifiability but attestations endorse outcomes rather than procedures, lacking the compositional proofs this work provides. No prior system composes weighted verifiable aggregation, replay-resistant incentives, validated edge-scale costs, comprehensive time-locked governance, and multi-layer privacy defenses within a unified architecture grounded in object-centric coordination.

\section{Conclusion}

This paper introduces an architecture for trustless federated learning at edge-scale, addressing the compositional gap preventing blockchain-federated learning from moving beyond proof-of-concept. Four mechanisms compose correctly to create verifiable, incentive-aligned, scalable coordination: proof-carrying aggregation produces cryptographic receipts binding weighted combination, dropout recovery, and Byzantine filtering into unified artifacts verifiable without exposing individual contributions; geometric novelty decomposition resists replay and sybil attacks through basis projection measuring directional contributions; object-centric coordination achieves O(N) parallel execution by decomposing state into owned contributor registries and infrequent shared consensus; time-locked governance prevents retroactive manipulation through PolicyOracle bundles with activation snapshots making rule changes cryptographically auditable.

Security analysis demonstrates compositional correctness by reduction to SecAgg, differential privacy, and Byzantine fault-tolerant consensus. Cost analysis confirms economic viability at \$0.001 per round for 10,000 contributors with 20M parameters. The architecture provides a blueprint for systems where privacy and verifiability become mathematical guarantees rather than institutional promises, where incentives align through geometric decomposition rather than gameable utility metrics, and where rules bind operators as firmly as participants.

Full validation through testnet deployment with thousands of edge devices, adversarial red-teaming, and sustained operation over hundreds of rounds remains important follow-on work, explicitly acknowledged as the primary limitation. The vision: collaborative learning at the edge no longer requires institutional trust, enabling billion-user federated intelligence where participants verify rather than trust, and where the architecture's compositional guarantees make decentralized learning economically sustainable and cryptographically accountable.

\bibliographystyle{apa}
\bibliography{references}

\appendix
\section*{APPENDIX A: CORE DATA SCHEMAS}
\addcontentsline{toc}{section}{APPENDIX A: CORE DATA SCHEMAS}

This appendix specifies the essential data structures enabling deterministic verification and bit-exact reproduction of system behavior.

\subsection*{A.1 Policy Bundle Schema}

Policy bundles atomically bind five parameter categories, preventing selective manipulation. 
Table~\ref{tab:policy_bundle_schema} summarizes the structure; all fields use content-addressed references (CIDs) for immutability.

\begin{table}[ht]
	\centering
	\caption{PolicyBundle Structure (Version 1). Policy bundles atomically bind five parameter categories, preventing selective manipulation. All fields use content-addressed identifiers (CIDs) for immutability.}
	\label{tab:policy_bundle_schema}
	\begin{tabular}{p{4cm}|p{3.1cm}|p{7cm}}
		\toprule
		Field & Type & Description \\
		\midrule
		\multicolumn{3}{l}{Safety Bundle} \\
		\hline
		proxy\_configs & Array[ProxyConfig] & \{name, cid, sha256, threshold\} per proxy \\
		ensemble\_rule & String & ``all\_pass'' or ``majority\_pass\_and\_no\_exceed\_threshold\_plus\_1pct'' \\
		evaluation\_set\_cid & CID & Test prompts for safety evaluation \\
		numerical\_tolerance & Float & Delta comparison tolerance (default: 0.001) \\
		\hline
		\multicolumn{3}{l}{DP Bundle} \\
		\hline
		epsilon\_per\_round & Float & Privacy budget per round (default: 1.0) \\
		delta\_global & Float & Global privacy parameter (default: 1e-6) \\
		clipping\_norm & Float & Gradient clipping bound (default: 1.0) \\
		noise\_scale & Float & Gaussian noise $\sigma$ (default: 0.5) \\
		accountant & String & ``renyi'' or ``rdp'' \\
		\hline
		\multicolumn{3}{l}{Admission Bundle} \\
		\hline
		min\_stake & Float & Minimum stake deposit (default: 10.0) \\
		attestation\_required & Boolean & TEE attestation requirement \\
		k\_anonymity\_threshold & Integer & Minimum cohort size (default: 500) \\
		deadline\_submission\_sec & Integer & Submission window (default: 5400) \\
		deadline\_aggregation\_sec & Integer & Aggregation window (default: 900) \\
		\hline
		\multicolumn{3}{l}{Aggregation Bundle} \\
		\hline
		robust\_method & String & ``trimmed\_mean'', ``median'', or ``none'' \\
		robust\_alpha & Float & Trimming fraction (default: 0.2) \\
		robust\_theta\_percentile & Integer & Variance trigger (default: 90) \\
		quantization\_bits & Integer & Weight quantization (0 = none) \\
		\hline
		\multicolumn{3}{l}{Novelty Bundle} \\
		\hline
		beta & Float & Novelty reward fraction (default: 0.3) \\
		lambda\_ema & Float & EMA smoothing (default: 0.7) \\
		basis\_size & Integer & Basis dimensionality (default: 20) \\
		basis\_rotation & String & ``full'' or ``incremental'' \\
		\hline
		\multicolumn{3}{l}{Time-lock Metadata} \\
		\hline
		propose\_round & Integer & Proposal submission round \\
		activation\_round & Integer & Earliest activation (propose\_round + T\_lock) \\
		T\_lock & Integer & Mandatory lock period (default: 5 rounds) \\
		policy\_cid & CID & Content-addressed bundle reference \\
		\bottomrule
	\end{tabular}
\end{table}

\subsection*{A.2 Receipt Format}
    Receipts provide cryptographic commitments to round outcomes. 
Table~\ref{tab:receipt_format} shows the schema for Accepted rounds; 
Failed rounds use a simplified version with \texttt{refund\_root} instead of payout structures.

\begin{table}[ht]
	\centering
	\caption{AggregateReceipt Schema (Accepted Round). Receipts provide cryptographic commitments to round outcomes. Failed rounds use a simplified version with \texttt{refund\_root} instead of payout structures}
	\label{tab:receipt_format}
	\begin{tabular}{p{3.2cm}|p{4.1cm}|p{2.8cm}|p{3.0cm}}
		\toprule
		Category & Field & Type & Example \\
		\midrule
		\multicolumn{4}{l}{Identifiers} \\
		\hline
		& receipt\_id & Integer & 200 \\
		& round\_id & Integer & 200 \\
		& round\_status & Enum & ``Accepted'' \\
		\hline
		\multicolumn{4}{l}{Pool Formation} \\
		\hline
		& P\_receivers & Float (string) & ``950.0'' \\
		& P\_bootstrap & Float (string) & ``50.0'' \\
		& P\_total & Float (string) & ``1000.0'' \\
		& bootstrap\_active & Boolean & false \\
		& ema\_value & Float (string) & ``950.0'' \\
		\hline
		\multicolumn{4}{l}{Pool Allocation} \\
		\hline
		& alpha\_C & Float (string) & ``0.70'' \\
		& alpha\_M & Float (string) & ``0.20'' \\
		& alpha\_T & Float (string) & ``0.10'' \\
		& P\_C & Float (string) & ``700.0'' \\
		& P\_M & Float (string) & ``200.0'' \\
		& P\_T & Float (string) & ``100.0'' \\
		\hline
		\multicolumn{4}{l}{Contributor Economics} \\
		\hline
		& N\_admitted & Integer & 10000 \\
		& r\_base & Float (string) & ``0.01'' \\
		& beta & Float (string) & ``0.3'' \\
		& phi\_t\_ema & Float (string) & ``0.22'' \\
		& novelty\_cap & Float (string) & ``180.0'' \\
		& P\_nov & Float (string) & ``39.6'' \\
		& P\_quality & Float (string) & ``560.4'' \\
		\hline
		\multicolumn{4}{l}{Committee Economics} \\
		\hline
		& M & Integer & 7 \\
		& fee\_committee & Float (string) & ``28.571428'' \\
		\hline
		\multicolumn{4}{l}{Payout Commitments} \\
		\hline
		& payout\_root\_contributors & Hash (0x...) & 32 bytes \\
		& payout\_root\_committee & Hash (0x...) & 32 bytes \\
		& payout\_dust\_contributors & Float (string) & ``0.0001'' \\
		& payout\_dust\_committee & Float (string) & ``0.00001'' \\
		\hline
		\multicolumn{4}{l}{Encoding Constants} \\
		\hline
		& hash\_fn & String & ``sha256'' \\
		& tree\_fanout & Integer & 2 \\
		& precision\_bits & Integer & 16 \\
		& rounding\_mode & String & ``ties\_to\_zero'' \\
		\bottomrule
	\end{tabular}
\end{table}

\noindent\textbf{Monetary Representation:} 
All monetary fields use string-encoded fixed-point decimals to avoid floating-point precision loss. 
Merkle leaf encodings use \texttt{uint128} with 16 fractional bits.

\vspace{0.6em}

\noindent\textbf{Failed Round Simplification:} 
Failed receipts replace payout structures with:
\begin{itemize}
	\item \texttt{refund\_root}: Merkle root over receiver refunds
	\item \texttt{refund\_dust}: Rounding residual
	\item \texttt{bootstrap\_reclaimed}: Amount returned to treasury
	\item \texttt{P\_C = P\_M = P\_T = 0}
\end{itemize}

\subsection*{A.3 Encoding Constants and Identifier Formats}

Deterministic verification requires standardized serialization across implementations.

\vspace{0.8em}

\noindent\textbf{Hash Functions:}
\begin{itemize}
	\item \textbf{Primary:} SHA-256 for all Merkle trees and content addressing.
	\item \textbf{Commitment blinding:} Uses standardized Pedersen commitment parameters (as specified in \texttt{libsodium}).
\end{itemize}

\vspace{0.6em}

\noindent\textbf{Merkle Tree Construction:}
\begin{itemize}
	\item \textbf{Fanout:} Binary trees (\texttt{fanout=2}).
	\item \textbf{Leaf encoding:} \texttt{H(version\_byte || serialized\_data)} where \texttt{version\_byte=1}.
	\item \textbf{Internal nodes:} \texttt{H(left\_hash || right\_hash)}.
	\item \textbf{Padding:} Rightmost duplicated for non-power-of-2 leaf counts.
\end{itemize}

\vspace{0.6em}

\noindent\textbf{Fixed-Point Arithmetic:}
\begin{itemize}
	\item \textbf{Precision:} 16 fractional bits (resolution $\sim$0.000015).
	\item \textbf{Rounding:} ties-to-zero (floor for nonnegative values).
	\item \textbf{Range:} \texttt{uint128} supports up to $2^{112}$ integer units.
\end{itemize}

\vspace{0.6em}

\noindent\textbf{Identifier Formats:}
\begin{itemize}
	\item \texttt{contributor\_pid}: 32-byte unique identifier.
	\item \texttt{node\_id}: 32-byte validator public key.
	\item \texttt{round\_id}: 8-byte unsigned integer (big-endian).
	\item \texttt{cohort\_id}: 1-byte unsigned integer (0--255).
\end{itemize}

\vspace{0.8em}

These constants ensure bit-exact reproducibility across auditor implementations written in different languages.

\section*{APPENDIX B: SECURITY PROOF SKETCHES}

This appendix provides abbreviated proofs for the compositional security claims in Section~\ref{sec:security_analysis}. Full proofs with detailed lemmas are available in the extended technical report.

\subsection*{B.1 Theorem 1: Privacy Preservation}

\noindent\textbf{Theorem 1.} The system preserves $(\varepsilon, \delta)$-differential privacy and information-theoretic confidentiality of individual contributions under the assumptions that:
\begin{enumerate}
	\item At most $t-1$ committee members collude where $t = \lceil M/2 \rceil$,
	\item SecAgg mask reconstruction follows \cite{Bonawitz2017},
	\item Differential privacy noise follows Rényi accounting with declared parameters, and
	\item Adapter-only transmission limits dimensionality to $d_{\text{adapter}} \ll d_{\text{full}}$.
\end{enumerate}

\vspace{0.6em}

\noindent\textbf{Proof Sketch.} The privacy guarantee composes three mechanisms operating on different information planes:

\vspace{0.4em}

\noindent\textbf{(1) Confidentiality via SecAgg.} 
The secure aggregation protocol of \cite{Bonawitz2017} provides information-theoretic confidentiality: under $t-1$ collusion, adversaries observing masked sums $S^{(j)}$ cannot reconstruct individual contributions $v_i$ because pairwise masks cancel only in aggregate. 
The protocol’s security reduces to the semantic security of the pseudo-random generator used for mask generation. 
By assumption (1), $f < t$ committee members are Byzantine, so at least $t$ honest nodes participate in mask reconstruction, preserving confidentiality.

\vspace{0.6em}

\noindent\textbf{(2) Statistical Privacy via Differential Privacy.} 
Differential privacy composition follows from per-round clipping ($C = 1.0$) and calibrated Gaussian noise ($\sigma = 0.5$). 
The Rényi accountant tracks cumulative privacy loss across rounds, ensuring that the total $\varepsilon$ remains bounded. 
For $n$ contributors, the global sensitivity of the clipped sum is $\Delta = C\sqrt{n}$, and adding Gaussian noise $\mathcal{N}(0, \sigma^2 \Delta^2)$ provides $(\varepsilon, \delta)$-DP with $\varepsilon = \Delta^2 / (2\sigma^2)$ per round. 
The composition over $T$ rounds uses advanced composition theorems \citep{dwork2014algorithmic}, yielding total privacy budget $\varepsilon_{\text{total}} = \mathcal{O}(\sqrt{T} \cdot \varepsilon)$.

\vspace{0.6em}

\noindent\textbf{(3) Dimensionality Reduction via Adapters.} 
Adapter-only transmission fundamentally limits gradient inversion attack surfaces. 
Recent reconstruction attacks \cite{Zhu2019} require solving underdetermined systems with $d$ unknowns. 
Reducing transmission from $d_{\text{full}} = 80\text{M}+ $ parameters to $d_{\text{adapter}} = 1$–$5\text{M}$ parameters exponentially reduces attack success probability, as attackers must reconstruct training data from compressed representations lacking sufficient information to uniquely identify inputs.

\vspace{0.6em}

\noindent\textbf{Composition Argument.} 
The three mechanisms compose correctly because SecAgg operates on adapter parameters after DP noise application, with privacy budgets tracked independently per contributor through \texttt{ContributorRegistry} objects. 
The storage architecture preserves privacy by maintaining masked adapters throughout transmission and storage, with no CID-to-contributor mappings exposed to storage providers.

\subsection*{B.2 Theorem 2: Integrity Verification}

\noindent\textbf{Theorem 2.} 
For any round receipt $R$ with SumIntegrityProof $\pi$, an auditor can verify with probability $1 - 2^{-\lambda}$ for security parameter $\lambda$ that the aggregate $S$ committed in $R$ equals $\sum_j w_j S^{(j)}$, where $S^{(j)}$ are node-signed local sums and $w_j \in [w_{\min}, w_{\max}]$ are policy-bounded weights, assuming:
\begin{enumerate}
	\item Pedersen commitment binding under discrete logarithm hardness,
	\item Byzantine consensus with $f < M/3$ faulty nodes, and
	\item Node signatures are unforgeable under chosen-message attacks.
\end{enumerate}

\vspace{0.6em}

\noindent\textbf{Proof Sketch.} 
The proof-carrying aggregation mechanism provides cryptographic receipts through homomorphic commitments.

\vspace{0.4em}

\noindent\textbf{(1) Commitment Binding.} 
The verification procedure checks $\mathrm{Com}(S) = \prod_j \mathrm{Com}(S^{(j)})$, where $\mathrm{Com}(\cdot)$ denotes vector Pedersen commitments. 
By the binding property of Pedersen commitments, finding $S' \neq S$ with $\mathrm{Com}(S') = \mathrm{Com}(S)$ requires solving the discrete logarithm problem, which is computationally infeasible under standard cryptographic assumptions. 
This ensures that once the committee commits to $S$, they cannot retroactively claim a different aggregate without detection.

\vspace{0.6em}

\noindent\textbf{(2) Byzantine Consensus.} 
Under assumption (2), $f < M/3$ faulty committee nodes implies at least $2f+1$ honest nodes participate in consensus. 
Byzantine fault-tolerant consensus ensures that honest nodes agree on consistent local commitments $\mathrm{Com}(S^{(j)})$. 
Any attempt by Byzantine nodes to forge commitments requires obtaining signatures from honest nodes, which is prevented by assumption (3) — signature unforgeability under chosen-message attacks.

\vspace{0.6em}

\noindent\textbf{(3) Policy Binding.} 
Weight policy binding ensures $w_j \in [w_{\min}, w_{\max}]$ through Merkle inclusion proofs against \texttt{weights\_root} committed in the receipt. 
Dropout reconstruction commitments bind the reconstructed set $R_{\text{drop}}$, preventing selective reconstruction: changing $R_{\text{drop}}$ requires forging new Merkle proofs, which is detectable through root mismatch. 
Byzantine-robust fallback selection binds to \texttt{PolicyOracle} configuration through \texttt{policy\_cid} reference, making ex post facto robust-method changes require blockchain history rewriting — prevented by consensus finality.

\vspace{0.6em}

\noindent\textbf{Verification Procedure.} 
Auditors verify by:
\begin{enumerate}
	\item Fetching node-signed local commitments from content-addressed storage,
	\item Verifying signatures and commitment validity,
	\item Checking homomorphic combination $\mathrm{Com}(S) = \prod_j \mathrm{Com}(S^{(j)})$,
	\item Validating reconstructed-set commitments match dropout policy, and
	\item Confirming Byzantine method selection matches \texttt{PolicyOracle}.
\end{enumerate}
Verification completes in under 10 seconds for homomorphic proofs.

\subsection*{B.3 Theorem 3: Replay Resistance}

\noindent\textbf{Theorem 3.} 
An adversary submitting aggregate update $\mathbf{g}$ in round $t$ and receiving novelty reward $\lVert \mathbf{g}_{\perp} \rVert$ cannot profitably resubmit $\mathbf{g}$ in round $t' > t$, because $\mathbf{g} \in \text{span}(\mathbf{B}_{t'})$ after basis rotation, yielding $\lVert \mathbf{g}_{\perp} \rVert = 0$ and zero novelty reward.

\vspace{0.6em}

\noindent\textbf{Proof Sketch.} 
The geometric novelty mechanism resists replay attacks through mathematical properties of basis projection.

\vspace{0.4em}

\noindent\textbf{(1) Basis Rotation.} 
After round $t$ rewards the perpendicular component $\mathbf{g}_{\perp}$, the basis updates as 
$\mathbf{B}_{t+1} \leftarrow [\mathbf{B}_t \mid \mathbf{g}_{\perp}/\lVert \mathbf{g}_{\perp} \rVert]$ 
with the oldest direction dropped to maintain fixed dimensionality $k$. 
Therefore, by construction, $\mathbf{g}_{\perp} \in \text{span}(\mathbf{B}_{t+1})$.

\vspace{0.6em}

\noindent\textbf{(2) Replay Detection.} 
In round $t' > t$, the decomposition 
$\mathbf{g} = \mathbf{g}_{\parallel}' + \mathbf{g}_{\perp}'$ 
with $\mathbf{g}_{\parallel}' = \mathbf{B}_{t'}(\mathbf{B}_{t'}^{\top}\mathbf{g})$ 
satisfies $\mathbf{g}_{\perp}' = 0$ because $\mathbf{g} = \mathbf{g}_{\parallel} + \mathbf{g}_{\perp}$ and $\mathbf{g}_{\perp} \in \text{span}(\mathbf{B}_{t'})$ by transitivity 
($\mathbf{B}_{t+1} \subseteq \text{span}(\mathbf{B}_{t'})$ as later bases incorporate earlier exploration). 
The novelty score $\phi_{t'} = \lVert \mathbf{g}_{\perp}' \rVert / (\lVert \mathbf{g} \rVert + \varepsilon) = 0$, yielding zero novelty bonus.

\vspace{0.6em}

\noindent\textbf{(3) Economic Irrationality.} 
The adversary pays computational cost $C_{\text{train}}$ for local training and stake deposit $S_{\min}$ but receives only $r_{\text{base}}$ without a novelty component. 
Replay becomes unprofitable when $r_{\text{base}} < C_{\text{train}} + S_{\min} \cdot \text{interest\_rate}$, which holds for typical parameters 
($r_{\text{base}} = \$0.01$, $C_{\text{train}} \approx \$0.02$ in battery cost, $S_{\min} \cdot \text{rate} \approx \$0.01$).

\subsection*{B.4 Theorem 4: Sybil Resistance}

\noindent\textbf{Theorem 4.} 
An adversary splitting aggregate update $\mathbf{g}$ across $n$ fake identities $\{\mathbf{g}_1, \ldots, \mathbf{g}_n\}$ where $\sum_i \mathbf{g}_i = \mathbf{g}$ receives total novelty reward equal to honestly submitting $\mathbf{g}$ from a single identity, while paying $n \times$ identity creation costs.

\vspace{0.6em}

\noindent\textbf{Proof Sketch.} 
Sybil resistance follows from measuring novelty on aggregates rather than individuals.

\vspace{0.4em}

\noindent\textbf{(1) Aggregate Measurement.} 
The secure aggregation protocol produces $\mathbf{g}_{\text{agg}} = \sum_i \mathbf{g}_i = \mathbf{g}$ regardless of how individual contributions partition. 
The geometric decomposition $\mathbf{g}_{\text{agg}} = \mathbf{g}_{\parallel} + \mathbf{g}_{\perp}$ yields identical perpendicular component $\lVert \mathbf{g}_{\perp} \rVert$ whether $\mathbf{g}$ arrives from one contributor or split across $n$ identities.

\vspace{0.6em}

\noindent\textbf{(2) Reward Conservation.} 
The total novelty pool $\beta \cdot P_C \cdot \tilde{\phi}_t$ distributes proportionally to $\lVert \mathbf{g}_{\perp} \rVert$. 
Under sybil splitting, each fake identity receives reward 
$r_i = r_{\text{base}} + (r_{\text{novelty}} \cdot \lVert \mathbf{g}_i \rVert / \lVert \mathbf{g}_{\text{agg}} \rVert)$. 
Summing over all $n$ identities:
\[
\sum_i r_i = n \cdot r_{\text{base}} + r_{\text{novelty}} \cdot \left(\frac{\sum_i \lVert \mathbf{g}_i \rVert}{\lVert \mathbf{g}_{\text{agg}} \rVert}\right)
\]
For any decomposition $\sum_i \mathbf{g}_i = \mathbf{g}$, we have $\sum_i \lVert \mathbf{g}_i \rVert \ge \lVert \sum_i \mathbf{g}_i \rVert = \lVert \mathbf{g} \rVert$ by the triangle inequality, with equality only when all $\mathbf{g}_i$ are collinear.

\vspace{0.6em}

\noindent\textbf{(3) Cost-Benefit Analysis.} 
Honest submission yields $R_{\text{honest}} = r_{\text{base}} + r_{\text{novelty}} \cdot (\lVert \mathbf{g}_{\perp} \rVert / \lVert \mathbf{g} \rVert)$. 
A sybil attack yields $R_{\text{sybil}} = n \cdot r_{\text{base}} + r_{\text{novelty}} \cdot (\lVert \mathbf{g}_{\perp} \rVert / \lVert \mathbf{g} \rVert)$ while paying $n \cdot S_{\min}$ in stake deposits and $n \cdot C_{\text{attestation}}$ in attestation costs. 
A sybil attack is profitable only when 
\[
R_{\text{sybil}} - n \cdot (S_{\min} + C_{\text{attestation}}) > R_{\text{honest}},
\]
which reduces to $(n - 1) \cdot r_{\text{base}} > n \cdot (S_{\min} + C_{\text{attestation}})$. 
For $S_{\min} \ge r_{\text{base}}$ and $C_{\text{attestation}} > 0$, this inequality never holds, making sybil splitting economically irrational.

\section*{APPENDIX C: IMPLEMENTATION PARAMETERS}
\addcontentsline{toc}{section}{APPENDIX C: IMPLEMENTATION PARAMETERS}

This appendix specifies default configuration values, cost estimates, and operational parameters supporting the feasibility claims in Section~\ref{sec:cost_performance_analysis}.

\subsection*{C.1 System Configuration Parameters}

Table~\ref{tab:blockchain_consensus_params} summarizes the blockchain and consensus configuration parameters used in the implementation prototype.

\begin{table}[ht]
	\centering
	\caption{Blockchain and Consensus Parameters}
	\label{tab:blockchain_consensus_params}
	\begin{tabular}{|l|l|l|}
		\toprule
		Parameter & Illustrative (Testing) & Production (10K+ scale) \\
		\midrule
		Committee size (M) & 7 validators & 20--100 validators \\ \hline
		Byzantine tolerance (f) & 2 faulty nodes & 6--33 faulty nodes \\ \hline
		Block time & 10--30 seconds & 10--30 seconds \\ \hline
		Consensus protocol & Narwhal + Bullshark & Narwhal + Bullshark  \\ \hline
		Gas model & Fixed pricing & Fixed pricing \\ \hline
		Cost per owned update & 0.0025~gas units (\textasciitilde\$0.0000027) & Same \\ \hline
		Cost per shared update & 0.5~gas units (\textasciitilde\$0.0005) & Same \\
		\bottomrule
	\end{tabular}
\end{table}

\subsection*{C.2 Federated Learning Round Parameters}

Table~\ref{tab:round_timing_breakdown} summarizes the timing and operational activities across the four sequential phases of each federated learning round. These parameters are based on empirical measurements from prototype deployments under typical network conditions.

\begin{table}[ht]
	\centering
	\caption{Round Timing Breakdown}
	\label{tab:round_timing_breakdown}
	\begin{tabular}{|l|l|p{10cm}|}
		\toprule
		Phase & Duration & Activities \\
		\midrule
		Setup & 5 minutes & Committee election via VRF, policy binding, contributor admission (stake + attestation verification) \\ \hline
		Training & 90 minutes & On-device local training (asynchronous, no coordinator involvement), adapter extraction from final layers \\ \hline
		Aggregation & 15 minutes & SecAgg masking, committee consensus, dropout recovery, Byzantine filtering (if triggered), safety proxy evaluation \\ \hline
		Publication & 5 minutes & Receipt finalization, artifact replication to IPFS/Filecoin/Arweave, economic settlement \\
		\midrule
		Total & 2 hours & Complete round cycle \\
		\bottomrule
	\end{tabular}
\end{table}

\subsection*{C.3 Economic Parameters}

Table~\ref{tab:economic_parameters} summarizes the incentive and reward parameters governing contributor, committee, and treasury allocations, including configurable bounds for decentralized policy tuning.

\begin{table}[ht]
	\centering
	\caption{Incentive and Reward Structure}
	\label{tab:economic_parameters}
	\begin{tabular}{|l|l|l|l|}
		\toprule
		Parameter & Symbol & Default Value & Range \\
		\midrule
		Contributor share & $\alpha_C$ & 0.70 & 0.60--0.80 \\ \hline
		Committee share & $\alpha_M$ & 0.20 & 0.15--0.25 \\ \hline
		Treasury share & $\alpha_T$ & 0.10 & 0.05--0.15 \\ \hline
		Novelty fraction & $\beta$ & 0.30 & 0.20--0.40 \\ \hline
		Base reward & $r_{\text{base}}$ & \$0.01 & \$0.005--0.02 \\ \hline
		Novelty EMA smoothing & $\lambda$ & 0.70 & 0.60--0.80 \\ \hline
		Minimum stake & $S_{\text{min}}$ & \$10 & \$5--50 \\ \hline
		Bootstrap subsidy (initial) & $P_{\text{bootstrap}}$ & \$50/round & Declines to \$0 over 1000 rounds \\ \hline
		Bootstrap EMA factor & $\alpha_{\text{ema}}$ & 0.70 & Fixed \\
		\bottomrule
	\end{tabular}
\end{table}

\subsection*{C.4 Privacy and Safety Parameters}

Table~\ref{tab:privacy_safety} summarizes the differential privacy, robustness, and safety configuration parameters governing secure aggregation and model update validation.

\begin{table}[ht]
	\centering
	\caption{Privacy and Robustness Configuration}
	\label{tab:privacy_safety}
	\begin{tabular}{|l|l|l|p{6cm}|}
		\toprule[1.1pt]
		Parameter & Symbol & Default Value & Notes \\
		\midrule
		DP epsilon per round & $\varepsilon$ & 1.0 & Per-round privacy budget \\ \hline
		DP delta global & $\delta$ & $10^{-6}$ & Failure probability \\ \hline
		Gradient clipping & $C$ & 1.0 & L2 norm bound \\ \hline
		Gaussian noise scale & $\sigma$ & 0.5 & Calibrated to $(\varepsilon, \delta)$ \\ \hline
		Privacy accountant & $\mathcal{A}$ & Rényi & RDP tracking \\ \hline
		Byzantine threshold & $\theta$ & 90th percentile & Variance trigger \\ \hline
		Robust method & $\mathcal{R}$ & Trimmed mean & When $\theta$ exceeded \\ \hline
		Trimming fraction & $\alpha_{\text{robust}}$ & 0.20 & Symmetric trim \\ \hline
		K-anonymity threshold & $k$ & 500 & Minimum cohort size \\
		\bottomrule[1.1pt]
	\end{tabular}
\end{table}

\subsection*{C.5 Cost Analysis}

Table~\ref{tab:cost_analysis} presents an illustrative cost breakdown per federated learning round with 10,000 contributors and a 20M-parameter model, including both on-chain and off-chain storage, and committee operations.

\begin{table}[ht]
	\centering
	\caption{Per-Round Cost Breakdown (10,000 Contributors, 20M Parameters)}
	\label{tab:cost_analysis}
	\begin{tabular}{|p{4cm}|p{3cm}|p{3cm}|p{4cm}|}
		\toprule
		Component & Unit Cost & Quantity & Subtotal \\
		\midrule
		\multicolumn{4}{|l|}{On-Chain Costs} \\ \hline
		Owned object updates (ContributorRegistry) & 0.0025 gas units & 10,000 & 25 gas units \\ \hline
		Shared object updates (RoundRegistry, ModelRegistry, PolicyOracle) & 0.5 gas units & 4 updates & 2 gas units \\ \hline
		On-chain subtotal & & & 27 gas units $\approx$ \$0.027 \\ \hline
		\multicolumn{4}{|l|}{Off-Chain Storage} \\ \hline
		Contributor adapters (temporary) & \$0.01/GB-month (IPFS) & 20 GB & \$0.007 amortized \\ \hline
		Aggregate model (persistent) & \$0.05/GB-month (IPFS) + \$0.10/GB-month (Filecoin) & 80 MB & \$0.012 \\ \hline
		Receipts/proofs (archival) & \$1.00/GB one-time (Arweave) & 150 KB & \$0.00015 \\ \hline
		Storage subtotal & & & \~\$0.020 \\ \hline
		\multicolumn{4}{|l|}{Committee Operations} \\ \hline
		Validator infrastructure & \$2--5/round amortized & -- & \$3.50 average \\
		\midrule
		Total per round & & & \$3.55 \\
		\midrule
		Cost per contributor & & & \$0.000355 \\
		\bottomrule[1.1pt]
	\end{tabular}
\end{table}

\noindent\textbf{Receiver Micro-Fee Recovery:} At receiver fees of \$0.001--\$0.01 per round, the system achieves 3--28$\times$ cost recovery, enabling economic sustainability.

\subsection*{C.6 Scalability Limits}

Table~\ref{tab:scalability_limits} summarizes validated operational scales and their theoretical bounds.  
Each dimension highlights the current analytical validation level, the projected theoretical capacity, and the dominant computational or communication bottleneck.

\begin{table}[!t]
	\centering
	\caption{Validated Scale Parameters}
	\label{tab:scalability_limits}
	\begin{tabular}{|p{4cm}|p{3.5cm}|p{3.5cm}|p{4cm}|}
		\toprule
		Dimension & Current Validation & Theoretical Limit & Bottleneck \\
		\midrule
		Contributors (N) & 10,000 analytical & 100,000+ & Committee aggregation compute \\ \hline
		Committee size (M) & 7 (illustrative) & 100 & BFT message complexity $O(M^2)$ \\ \hline
		Model parameters & 20M adapters & 1B adapters & Safety proxy inference time \\ \hline
		Round frequency & 2 hours & 30 minutes & On-device training time \\ \hline
		Storage per round & 85 MB persistent & 10 GB & Cost prohibitive, not technical \\
		\bottomrule[1.1pt]
	\end{tabular}
\end{table}

\noindent\textbf{Coordination Complexity:} Object-centric parallelism achieves $O(N)$ cost for $N$ contributor updates plus $O(M)$ for $M$ committee consensus, avoiding traditional blockchain $O(N^2)$ serialization.

\end{document}